\documentclass[letterpaper, 10 pt, conference]{ieeeconf}  

\IEEEoverridecommandlockouts                              

\usepackage{dblfloatfix} 
\usepackage[utf8]{inputenc} 
\usepackage[T1]{fontenc}    
\usepackage{url}            
\usepackage{booktabs}       
\usepackage{amsfonts}       
\usepackage{nicefrac}       
\usepackage{microtype}      
\usepackage{amsmath}
\usepackage{subcaption}
\usepackage{wrapfig}
\usepackage{bbold}
\usepackage{algorithm}
\usepackage{algorithmic}
\usepackage{xcolor}
\usepackage{hyperref}       
\hypersetup{
 colorlinks=True,
 linkcolor=blue,
 citecolor=blue,
 urlcolor=blue}

\usepackage[font=small,labelfont=bf]{caption}

\usepackage{graphics} 
\usepackage{epsfig} 
\usepackage{amssymb}  
\usepackage{wrapfig}
\usepackage{authblk}

\newcommand{\siteurl}{\url{https://sites.google.com/view/mtrf}}

\title{Reset-Free Reinforcement Learning via Multi-Task Learning: \\ Learning Dexterous Manipulation Behaviors without Human Intervention}

\author{Abhishek Gupta$^*$${^1}$\thanks{$^*$Authors contributed equally to this work.}
\hspace{3mm} Justin Yu$^*$${^1}$\thanks{~\siteurl}
\hspace{3mm} Tony Z. Zhao$^*$${^1}$
\hspace{3mm} Vikash Kumar$^*$${^2}$\\
\hspace{3mm} Aaron Rovinsky${^1}$
\hspace{3mm} Kelvin Xu${^1}$
\hspace{3mm} Thomas Devlin${^1}$
\hspace{3mm} Sergey Levine${^1}$\\
$^1$ UC Berkeley \hspace{0.2in} $^2$ University of Washington \hspace{0.2in}\\
}

\newcommand{\methodname}{\textit{MTRF}}

\newif\ifcomments
\commentstrue 
\ifcomments
    
    \providecommand{\tony}[2][]{{\protect\color{orange}{[Tony:\textbf{#1} #2]}}}
    \providecommand{\justin}[2][]{{\protect\color{violet}{[Justin:\textbf{#1} #2]}}}
    \providecommand{\vikash}[2][]{{\protect\color{blue}{[Vikash:\textbf{#1} #2]}}}

\else
    \providecommand{\eric}[2][]{}
    \providecommand{\tony}[2][]{}
    \providecommand{\justin}[2][]{}
    \providecommand{\vikash}[2][]{}
\fi

\begin{document}

\maketitle

\begin{abstract}
Reinforcement Learning (RL) algorithms can in principle acquire complex robotic skills by learning from large amounts of data in the real world, collected via trial and error. However, most RL algorithms use a carefully engineered setup in order to collect data, requiring human supervision and intervention to provide episodic resets. This is particularly evident in challenging robotics problems, such as dexterous manipulation. To make data collection scalable, such applications require reset-free algorithms that are able to learn autonomously, without explicit instrumentation or human intervention. Most prior work in this area handles single-task learning. However, we might also want robots that can perform large repertoires of skills. At first, this would appear to only make the problem harder. However, the key observation we make in this work is that an appropriately chosen multi-task RL setting actually alleviates the reset-free learning challenge, with minimal additional machinery required. In effect, solving a multi-task problem can directly solve the reset-free problem since different combinations of tasks can serve to perform resets for other tasks. By learning multiple tasks together and appropriately sequencing them, we can effectively learn all of the tasks together reset-free. This type of multi-task learning can effectively scale reset-free learning schemes to much more complex problems, as we demonstrate in our experiments. We propose a simple scheme for multi-task learning that tackles the reset-free learning problem, and show its effectiveness at learning to solve complex dexterous manipulation tasks in both hardware and simulation without any explicit resets. This work shows the ability to learn dexterous manipulation behaviors in the real world with RL without any human intervention.
\end{abstract}

\section{Introduction}

\begin{figure}[!h]
    \centering
        \includegraphics[width=0.95\columnwidth]{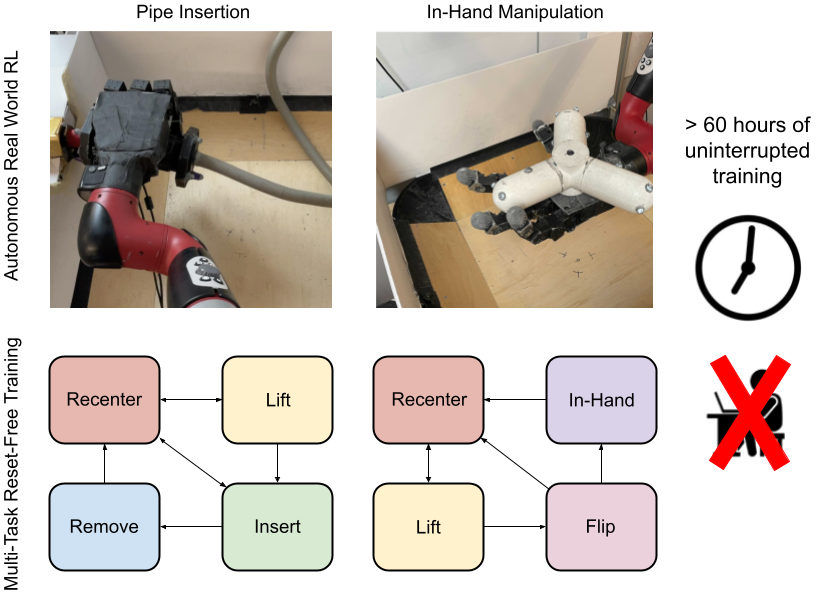}
    \caption{Reset-free learning of dexterous manipulation behaviors by leveraging multi-task learning. When multiple tasks are learned together, different tasks can serve to reset each other, allowing for uninterrupted continuous learning of all of the tasks. This allows for the learning of dexterous manipulation tasks like in-hand manipulation and pipe insertion with a 4-fingered robotic hand, without any human intervention, with over 60 hours of uninterrupted training}
    \label{fig:inhand-mainfig}
\end{figure}

Reinforcement learning algorithms have shown promise in enabling robotic tasks in simulation \cite{DAPG, 2018-TOG-deepMimic}, and even some tasks in the real world \cite{levine2015learning, pinto2016supersizing}. RL algorithms in principle can learn generalizable behaviors with minimal manual engineering, simply by collecting their own data via trial and error. This approach works particularly well in simulation, where data is cheap and abundant. Success in the real world has been restricted to settings where significant environment instrumentation and engineering is available to enable autonomous reward calculation and episodic resets \cite{andrychowicz2020learning, qtopt, pddm, r3l}. To fully realize the promise of robotic RL in the real world, we need to be able to learn even in the absence of environment instrumentation.

In this work, we focus specifically on reset-free learning for dexterous manipulation, which presents an especially clear lens on the reset-free learning problem. For instance, a dexterous hand performing in-hand manipulation, as shown in Figure~\ref{fig:inhand-mainfig} (right), must delicately balance the forces on the object to keep it in position. Early on in training, the policy will frequently drop the object, necessitating a particularly complex reset procedure. Prior work has addressed this manually by having a human involved in the training process~\cite{ploeger2020high, kumaroptimal, dppt}, instrumenting a reset in the environment~\cite{zhu2019dexterous, pi2gps}, or even by programming a separate robot to place the object back in the hand~\cite{pddm}. Though some prior techniques have sought to learn some form of a ``reset'' controller \cite{eysenbach2017leave, ahn2020robel, pulkit, r3l, asymmetricselfplay, richter17visual}, none of these are able to successfully scale to solve a complex dexterous manipulation problem without hand-designed reset systems due to the challenge of learning robust reset behaviors.

However, general-purpose robots deployed in real-world settings will likely be tasked with performing many different behaviors. While multi-task learning algorithms in these settings have typically been studied in the context of improving sample efficiency and generalization~\cite{distral, policydistillation, actor-mimic, pcgrad, ozansener, meta-world}, in this work we make the observation that multi-task algorithms naturally lend themselves to the reset-free learning problem.
We hypothesize that the reset-free RL problem can be addressed by reformulating it as a multi-task problem, and appropriately sequencing the tasks commanded and learned during online reinforcement learning. As outlined in Figure~\ref{fig:inhand-taskgraph}, solving a collection of tasks simultaneously presents the possibility of using some tasks as a reset for others, thereby removing the need of explicit per task resets. For instance, if we consider the problem of learning a variety of dexterous hand manipulation behaviors, such as in-hand reorientation, then learning and executing behaviors such as recenter and pickup can naturally reset the other tasks in the event of failure (as we describe in Section~\ref{sec:method} and Figure~\ref{fig:inhand-taskgraph}). We show that by learning multiple different tasks simultaneously and appropriately sequencing behavior across different tasks, we can learn \emph{all} of the tasks without episodic resets required at all. This allows us to effectively learn a ``network'' of reset behaviors, each of which is easier than learning a complete reset controller, but together can execute and learn more complex behavior.

The main contribution of this work is to propose a learning system that can learn dexterous manipulation behaviors without the need for episodic resets. We do so by leveraging the paradigm of multi-task reinforcement learning to make the reset free problem less challenging. The system accepts a diverse set of tasks that are to be learned, and then trains reset-free, leveraging progress in some tasks to provide resets for other tasks. To validate this algorithm for reset-free robotic learning, we perform both simulated and hardware experiments on a dexterous manipulation system with a four fingered anthropomorphic robotic hand. To our knowledge, these results demonstrate the first instance of a combined hand-arm system learning dexterous in-hand manipulation with deep RL entirely in the real world with minimal human intervention during training, simultaneously acquiring both the in-hand manipulation skill and the skills needed to retry the task. We also show the ability of this system to learn other dexterous manipulation behaviors like pipe insertion via uninterrupted real world training, as well as several tasks in simulation. 

\section{Related Work}
\vspace{-1mm}

RL algorithms have been applied to a variety of robotics problems in simulation~\cite{james2019rlbench, DAPG, deepmindheess, openAIhand}, and have also seen application to real-world problems, such as locomotion~\cite{ha2020learning,peng2020learning, calandra2016gait},  grasping~\cite{qtopt, handeye, baier07grasping, wu2019mat}, manipulation of articulated objects~\cite{nemec2017door, doorgym, pi2gps, ddpgshane, RIG}, and even dexterous manipulation~\cite{vanHoof2015learning, kumaroptimal}. Several prior works \cite{okamura2000overview} have shown how to acquire dexterous manipulation behaviors with optimization \cite{furukawa2006dynamic, bai2014dexterous, mordatch2012contact, mpcvikash, yamane2004synthesizing}, reinforcement learning in simulation \cite{DAPG, mandikal2020dexterous, jain2019learning}, and even in the real world \cite{dexterousopenai, ploeger2020high, vanhoof, pddm, ahn2020robel, zhu2019dexterous, kumaroptimal, softhand, choisofthand, kumar2016learning}. These techniques have leaned on highly instrumented setups to provide episodic resets and rewards. For instance, prior work uses a scripted second arm \cite{pddm} or separate servo motors \cite{zhu2019dexterous} to perform resets. Contrary to these, our work focuses on removing the need for explicit environment resets, by leveraging multi-task learning.

Our work is certainly not the first to consider the problem of reset-free RL~\cite{rfgps}. \cite{eysenbach2017leave} proposes a scheme that interleaves attempts at the task with episodes from an explicitly learned reset controller, trained to reach the initial state. Building on this work, \cite{r3l} shows how to learn simple dexterous manipulation tasks without instrumentation using a perturbation controller exploring for novelty instead of a reset controller. \cite{han2015learning, smith2019avid} demonstrate learning of multi-stage tasks by progressively sequencing a chain of forward and backward controllers. Perhaps the most closely related work to ours algorithmically is the framework proposed in ~\cite{ha2020learning}, where the agent learns locomotion by leveraging multi-task behavior. However, this work studies tasks with cyclic dependencies specifically tailored towards the locomotion domain. Our work shows that having a variety of tasks and learning them all together via multi-task RL can allow solutions to challenging reset free problems in dexterous manipulation domains.

Our work builds on the framework of multi-task learning~\cite{distral, policydistillation, actor-mimic, pcgrad, ozansener, meta-world, rudermtl, yang2020mtrl} which have been leveraged to learn a collection of behaviors, improve on generalization as well as sample efficiency, and have even applied to real world robotics problems~\cite{sacx, wulfmeiercompositional, deisenroth2014mtrl}. In this work, we take a different view on multi-task RL as a means to solve reset-free learning problems.

\section{Preliminaries}
\label{sec:prelims}

We build on the framework of Markov decision processes for reinforcement learning. We refer the reader to \cite{suttonandbarto} for a more detailed overview. RL algorithms consider the problem of learning a policy $\pi(a|s)$ such that the expected sum of rewards $R(s_t, a_t)$ obtained under such a policy is maximized when starting from an initial state distribution $\mu_0$ and dynamics $\mathcal{P}(s_{t+1}|s_t, a_t)$. This objective is given by:
\begin{align}\label{eq:standard-rl}
    J(\pi) = \mathbb{E}_{\substack{s_0 \sim \mu_0 \\ a_t \sim \pi(a_t|s_t) \\ s_{t+1} \sim \mathcal{P}(s_{t+1}| s_t, a_t)}}\left[\sum_{t=0}^T \gamma^t R(s_t, a_t)\right]
\end{align}

While many algorithms exist to optimize this objective, in this work we build on the framework of actor-critic algorithms~\cite{konda}. Although we build on actor critic framework, we emphasize that our framework can be effectively used with many standard reinforcement learning algorithms with minimal modifications.

As we note in the following section, we address the reset-free RL problem via multi-task RL. Multi-task RL attempts to learn multiple tasks simultaneously. Under this setting, each of $K$ tasks involves a separate reward function $R_i$, different initial state distribution $\mu_0^i$ and potentially different optimal policy $\pi_i$. Given a distribution over the tasks $p(i)$, the multi-task problem can then be described as
\begin{align}\label{eq:multi-task-rl}
    J(\pi_0, \ldots, \pi_{K-1}) = \mathbb{E}_{i \sim p(i)} \mathbb{E}_{\substack{s_0 \sim \mu_0^i \\ a_t \sim \pi_{i}(s_t|a_t) \\ s_{t+1} \sim \mathcal{P}(s_t, a_t)}}\left[\sum_t \gamma^t R_{i}(s_t, a_t)\right]
\end{align}

In the following section, we will discuss how viewing reset-free learning through the lens of multi-task learning can naturally address the challenges in reset-free RL.

\begin{figure*}[!ht]
    \centering
        \includegraphics[width=0.9\textwidth]{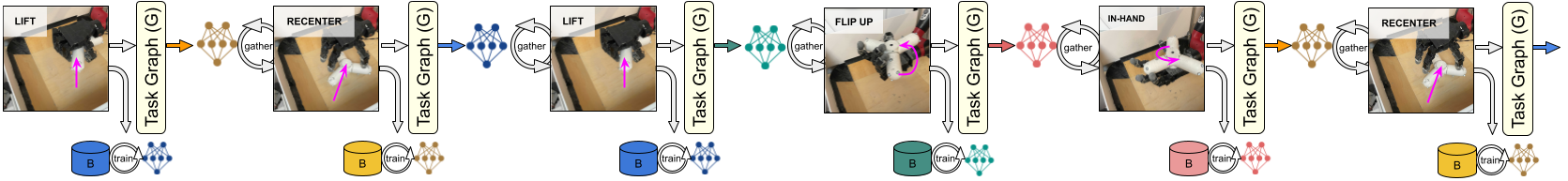}
    \caption{Depiction of some steps of reset-free training for the in-hand manipulation task family on hardware. Reset-free training uses the task graph to choose which policy to execute and train at every step. For executions that are not successful (e.g., the pickup in step 1), other tasks (recenter in step 2) serve to provide a reset so that pickup can be attempted again. Once pickup is successful, the next task (flip up) can be attempted. If the flip-up policy is successful, then the in-hand reorientation task can be attempted and if this drops the object then the re-centering task is activated to continue training.}
    \label{fig:inhand-trainflow}
\end{figure*}
\section{Learning Dexterous Manipulation Behaviors Reset-Free via Multi-Task RL}
\label{sec:method}

One of the main advantages of dexterous robots is their ability to perform a wide range of \emph{different} tasks. Indeed, we might imagine that a real-world dexterous robotic system deployed in a realistic environment, such as a home or office, would likely need to perform a repertoire of different behaviors, rather than just a single skill. While this may at first seem like it would only make the problem of learning without resets more difficult, the key observation we make in this work is that the multi-task setting can actually facilitate reset-free learning without manually provided instrumentation. When a large number of diverse tasks are being learned simultaneously, some tasks can naturally serve as resets for other tasks during learning. Learning each of the tasks individually without resets is made easier by appropriately learning and sequencing together other tasks in the right order. By doing so, we can replace the simple forward, reset behavior dichotomy with a more natural ``network'' of multiple tasks that can perform complex reset behaviors between each other.

Let us ground this intuition in a concrete example. Given a dexterous table-top manipulation task, shown in Fig~\ref{fig:inhand-taskgraph} and Fig~\ref{fig:inhand-trainflow} , our reset-free RL procedure might look like this: let us say the robot starts with the object in the palm and is trying to learn how to manipulate it in-hand so that it is oriented in a particular direction (\textit{in-hand reorient}). While doing so, it  may end up dropping the object. When learning with resets, a person would need to pick up the object and place it back in the hand to continue training. However, since we would like the robot to learn without such manual interventions, the robot itself needs to retrieve the object and resume practicing. To do so, the robot must first \textit{re-center} and the object so that it is suitable for grasping, and then actually \textit{lift} and the \textit{flip up} so it's in the palm to resume practicing. In case any of the intermediate tasks (say lifting the object) fails, the recenter task can be deployed to attempt picking up again, practicing these tasks themselves in the process. Appropriately sequencing the execution and learning of different tasks, allows for the autonomous practicing of in-hand manipulation behavior, without requiring any human or instrumented resets.

\subsection{Algorithm Description}
In this work, we directly leverage this insight to build a dexterous robotic system that learns in the absence of resets.  We assume that we are provided with $K$ different tasks that need to be learned together. These tasks each represent some distinct capability of the agent. As described above, in the dexterous manipulation domain the tasks might involve re-centering, picking up the object, and reorienting an object in-hand. Each of these $K$ different tasks is provided with its own reward function $R_i(s_t, a_t)$, and at test-time is evaluated against its distinct initial state distribution $\mu_0^i$.

Our proposed learning system, which we call Multi-Task learning for Reset-Free RL (\methodname), attempts to jointly learn $K$ different policies $\pi_i$, one for each of the defined tasks, by leveraging off-policy RL and continuous data collection in the environment. The system is initialized by randomly sampling a task and state $s_0$ sampled from the task's initial state distribution. The robot collects a continuous stream of data \textit{without any subsequent resets} in the environment by sequencing the $K$ policies according to a meta-controller (referred to as a ``task-graph'') $G(s) : S \rightarrow \{ 0, 1, \ldots, K - 1 \}$. Given the current state of the environment and the learning process, the task-graph makes a decision once every $T$ time steps on which of the tasks should be executed and trained for the \emph{next} $T$ time steps. This task-graph decides what order the tasks should be learned and which of the policies should be used for data collection. The learning proceeds by iteratively collecting data with a policy $\pi_i$ chosen by the task-graph for $T$ time steps, after which the collected data is saved to a task-specific replay buffer $\mathcal{B}_i$ and the task-graph is queried again for which task to execute next, and the whole process repeats.

We assume that, for each task, successful outcomes for tasks that lead into that task according to the task graph (i.e., all incoming edges) will result in valid initial states for that task. This assumption is reasonable: intuitively, it states that an edge from task A to task B implies that successful outcomes of task A are valid initial states for task B. This means that, if task B is triggered after task A, it will learn to succeed from these valid initial states under $\mu_0^B$. While this does not always guarantee that the downstream controller for task B will see \emph{all} of the initial states from $\mu_0^B$, since the upstream controller is not explicitly optimizing for coverage, in practice we find that this still performs very well. However, we expect that it would also be straightforward to introduce coverage into this method by utilizing state marginal matching methods~\cite{eysenbachsmm}. We leave this for future work.

The individual policies can continue to be trained by leveraging the data collected in their individual replay buffers $\mathcal{B}_i$ via off-policy RL. As individual tasks become more and more successful, they can start to serve as effective resets for other tasks, forming a natural curriculum. The proposed framework is general and capable of learning a diverse collection of tasks reset-free when provided with a task graph that leverages the diversity of tasks. This leaves open the question of how to actually define the task-graph $G$ to effectively sequence tasks. In this work, we assume that a task-graph defining the various tasks and the associated transitions is provided to the agent by the algorithm designer.  In practice, providing such a graph is simple for a human user, although it could in principle be learned from experiential data. We leave this extension for future work. Interestingly, many other reset-free learning methods~\cite{eysenbach2017leave, han2015learning, smith2019avid} can be seen as special cases of the framework we have described above. In our experiments we incorporate one of the tasks as a ``perturbation'' task. While prior work considered doing this with a single forward controller~\cite{r3l}, we show that this type of perturbation can generally be applied by simply viewing it as another task. We incorporate this perturbation task in our instantiation of the algorithm, but we do not show it in the task graph figures for simplicity.

\subsection{Practical Instantiation}
To instantiate the algorithmic idea described above as a deep reinforcement learning framework that is capable of solving dexterous manipulation tasks without resets, we can build on the framework of actor-critic algorithms. We learn separate policies $\pi_i$ for each of the $K$ provided tasks, with separate critics $Q_i$ and replay buffers $\mathcal{B}_i$ for each of the tasks. Each of the policies $\pi_i$ is a deep neural network Gaussian policy with parameters $\theta_i$, which is trained using a standard actor-critic algorithm, such as soft actor-critic \cite{haarnoja2018soft}, using data sampled from its own replay buffer $\mathcal{B}_i$. The task graph $G$ is represented as a user-provided state machine, as shown in Fig~\ref{fig:inhand-taskgraph}, and is queried every $T$ steps to determine which task policy $\pi_i$ to execute and update next. Training proceeds by starting execution from a particular state $s_0$ in the environment, querying the task-graph $G$ to determine which policy $i = G(s_0)$ to execute, and then collecting $T$ time-steps of data using the policy $\pi_i$, transitioning the environment to a new state $s_T$ (Fig \ref{fig:inhand-taskgraph}). The task-graph is then queried again and the process is repeated until all the tasks are learned. 

\begin{algorithm}[H]
  	\caption{\methodname}
  	\label{alg:method}
  	\begin{algorithmic}[1]
  	\STATE Given: $K$ tasks with rewards $R_i(s_t, a_t)$, along with a task graph mapping states to a task index $G(s) : S \rightarrow \{0, 1, \ldots, K - 1\}$
  	\STATE Let $\hat{i}$ represent the task index associated with the forward task that is being learned.
  	\STATE Initialize $\pi_i$, $Q_i$, $\mathcal{B}_i \text{ } \forall i \in \{0, 1, \ldots, K - 1\}$
  	\STATE Initialize the environment in task $\hat{i}$ with initial state $s_{\hat{i}} \sim \mu_{\hat{i}}(s_{\hat{i}})$
  	\FOR{iteration $n=1, 2, ...$}
  	    \STATE Obtain current task $i$ to execute by querying task graph at the current environment state $i = G(s_{\text{curr}})$
        \FOR{iteration $j=1, 2, ..., T$}
            \STATE Execute $\pi_i$ in environment, receiving task-specific rewards $R_i$ storing data in the buffer $\mathcal{B}_i$
            \STATE Train the current task's policy and value functions  $\pi_i$, $Q_i$ by sampling a batch from the replay buffer containing this task's experience $\mathcal{B}_i$, according to SAC \cite{haarnoja2018soft}.
        \ENDFOR
  	\ENDFOR
  	\end{algorithmic}
\end{algorithm}

\section{Task and System Setup}
\label{sec:setup}
\begin{figure}[!h]
    \centering
    \includegraphics[width=0.6\linewidth]{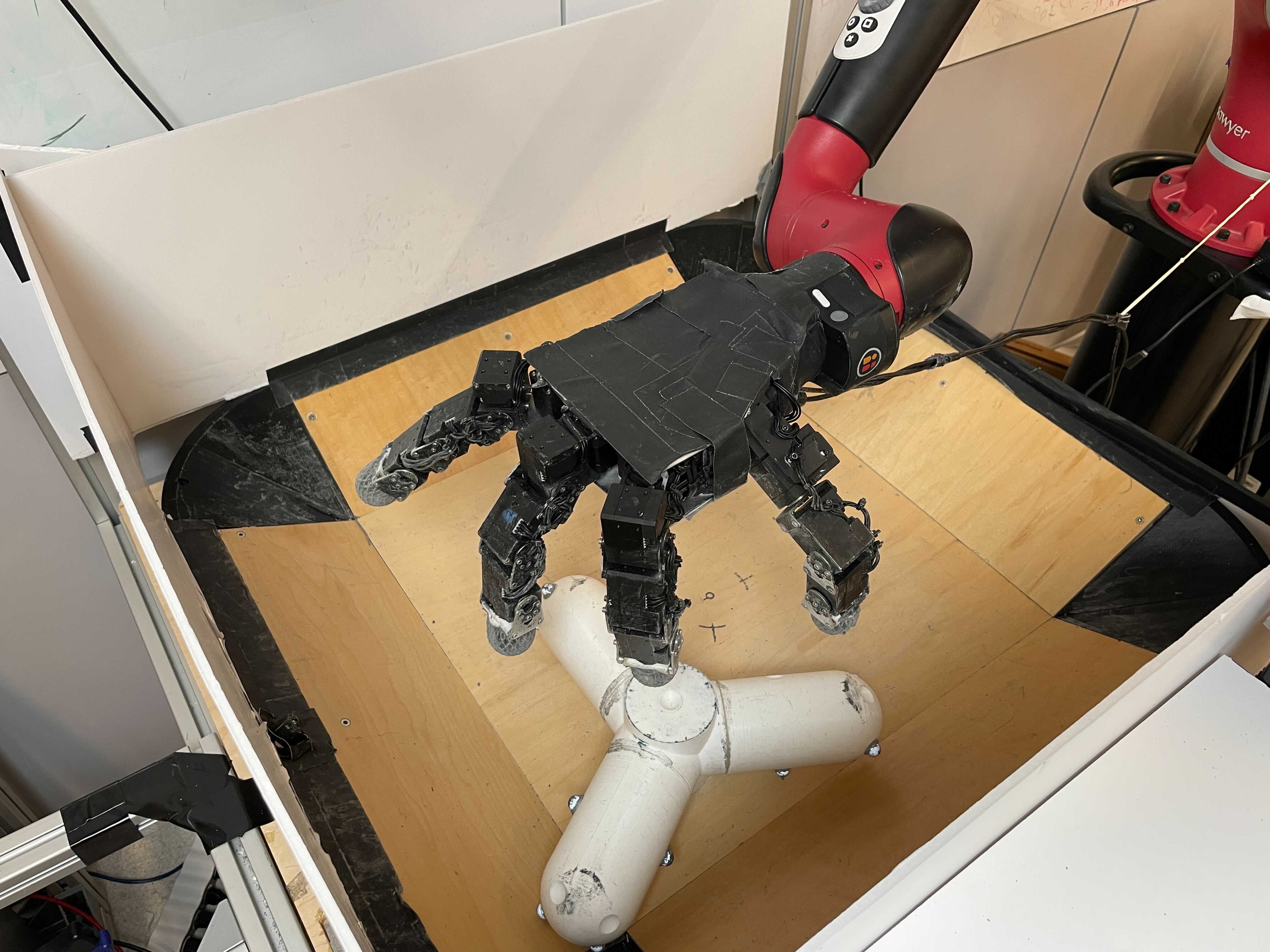}
    \caption{Real-world hand-arm manipulation platform. The system comprises a 16 DoF hand mounted on a 6 DoF Sawyer arm. The goal of the task is to perform in-hand reorientation, as illustrated in Fig~\ref{fig:example_trajectory_hardware_valve} or pipe insertion as shown in Fig~\ref{fig:example_trajectory_hardware_pipe}}
    \label{fig:real-setup}
\end{figure}

To study \methodname{} in the context of challenging robotic tasks, such as dexterous manipulation, we designed an anthropomorphic manipulation platform in both simulation and hardware. Our system (Fig~\ref{fig:real-setup}) consists of a 22-DoF anthropomorphic hand-arm system. We use a self-designed and manufactured four-fingered, 16 DoF robot hand called the \textit{D'Hand}, mounted on a 6 DoF Sawyer robotic arm to allow it to operate in an extended workspace in a table-top setting. We built this hardware to be particularly amenable to our problem setting due it's robustness and ease of long term operation. The \textit{D'Hand} can operate for upwards of $100$ hours in contact rich tasks without any breakages, whereas previous hand based systems are much more fragile. Given the modular nature of the hand, even if a particular joint malfunctions, it is quick to repair and continue training. In our experimental evaluation, we use two different sets of dexterous manipulation tasks in simulation and two different sets of tasks in the real world. Details can be found in Appendix A and at \siteurl 

\subsection{Simulation Domains}
\label{sec:sim-setup}

\begin{figure}[!h]
  \centering
  \includegraphics[width= 0.9\linewidth]{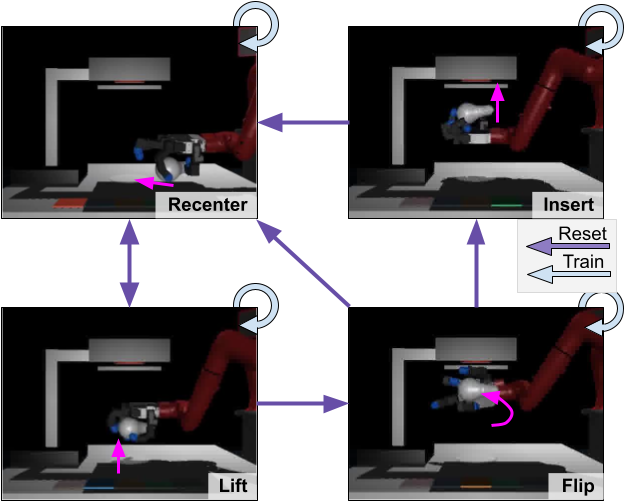}
  \caption{Tasks and transitions for lightbulb insertion in simulation. The goal is to recenter a lightbulb, lift it, flip it over, and then insert it into a lamp.}
  \label{fig:lightbulb-Domain}
\end{figure}

\noindent \textbf{Lightbulb insertion tasks.}
The first family of tasks involves inserting a lightbulb into a lamp in simulation with the dexterous hand-arm system. The tasks consist of centering the object on the table, pickup, in-hand reorientation, and insertion into the lamp. The multi-task transition task graph is shown in Fig \ref{fig:lightbulb-Domain}. These tasks all involve coordinated finger and arm motion and require precise movement to insert the lightbulb.

\begin{figure}[!h]
  \centering
  \includegraphics[width= 0.9\linewidth]{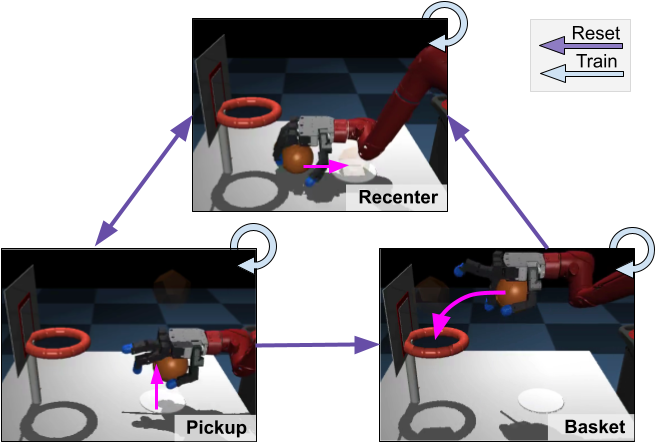}
  \caption{Tasks and transitions for basketball domain in simulation. The goal here is to reposition a basketball object, pick it up and then dunk it in a hoop.}
  \label{fig:Basketball-Domain}
\end{figure}

\noindent \textbf{Basketball tasks.}
The second family of tasks involves dunking a basketball into a hoop. This consists of repositioning the ball, picking it up, positioning the hand over the basket, and dunking the ball. This task has a natural cyclic nature, and allows tasks to reset each other as shown in Fig \ref{fig:Basketball-Domain}, while requiring fine-grained behavior to manipulate the ball midair.

\subsection{Hardware Tasks}
\label{sec:hardware-setup}

\begin{figure}[!h]
    \centering
        \includegraphics[width=0.9\columnwidth]{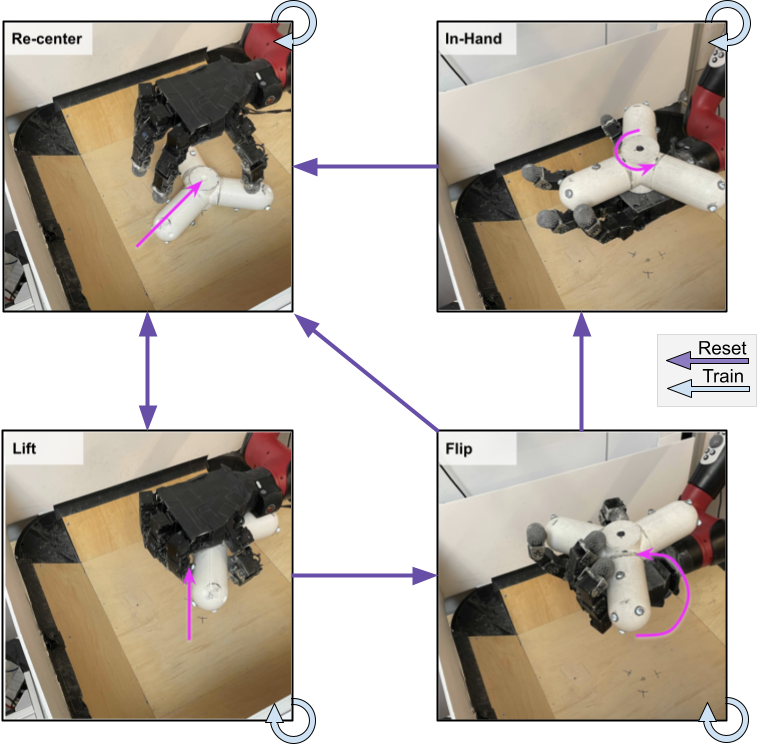}
    \caption{Tasks and transitions for the in-hand manipulation domain on hardware. The goal here is to rotate a 3 pronged valve object to a particular orientation in the palm of the hand, picking it up if it falls down to continue practicing.}
    \label{fig:inhand-taskgraph}
\end{figure}

We also evaluate \methodname{} on the real-world hand-arm robotic system, training a set of tasks in the real world, without any simulation or special instrumentation. We considered 2 different task families - in hand manipulation of a 3 pronged valve object, as well as pipe insertion of a cylindrical pipe into a hose attachment mounted on the wall. We describe each of these setups in detail below:

\textbf{In-Hand Manipulation:} For the first task on hardware, we use a variant of the in-hand reorienting task, where the goal is to pick up an object and reorient it in the palm into a desired configuration, as shown in Fig~\ref{fig:inhand-taskgraph}. This task not only requires mastering the contacts required for a successful pickup, but also fine-grained finger movements to reorient the object in the palm, while at the same time balancing it so as to avoid dropping. The task graph corresponding to this domain is shown in Fig~\ref{fig:inhand-taskgraph}. A frequent challenge in this domain stems from dropping the object during the in-hand reorientation, which ordinarily would require some sort of reset mechanism (as seen in prior work~\cite{pddm}). However, \methodname{} enables the robot to utilize such ``failures'' as an opportunity to practice the tabletop re-centering, pickup, and flip-up tasks, which serve to ``reset'' the object into a pose where the reorientation can be attempted again.\footnote{For the pickup task, the position of the arm's end-effector is scripted and only D'Hand controls are learned to reduce training time.} The configuration of the 22-DoF hand-arm system mirrors that in simulation. The object is tracked using a motion capture system. Our policy directly controls each joint of the hand and the position of the end-effector. The system is set up to allow for extended uninterrupted operation, allowing for over 60 hours of training without any human intervention. We show how our proposed technique allows the robot to learn this task in the following section. 

\begin{figure}[!h]
    \centering
        \includegraphics[width=0.9\columnwidth]{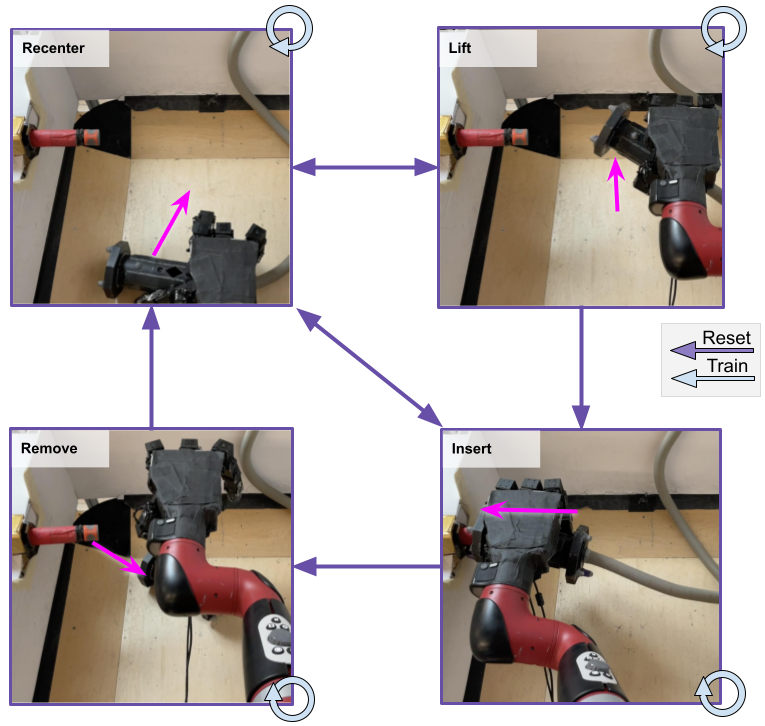}
    \caption{Tasks and transitions for pipe insertion domain on hardware. The goal here is to reposition a cylindrical pipe object, pick it up and then insert it into a hose attachment on the wall.}
    \label{fig:pipe-taskgraph}
\end{figure}

\textbf{Pipe insertion:} For the second task on hardware, we set up a pipe insertion task, where the goal is to pick up a cylindrical pipe object and insert it into a hose attachment on the wall, as shown in Fig~\ref{fig:pipe-taskgraph}. This task not only requires mastering the contacts required for a successful pickup, but also accurate and fine-grained arm motion to insert the pipe into the attachment in the wall. The task graph corresponding to this domain is shown in Fig~\ref{fig:pipe-taskgraph}. In this domain, the agent learns to pickup the object and then insert it into the attachment in the wall. If the object is dropped, it is then re-centered and picked up again to allow for another attempt at insertion. \footnote{For the pickup task, the position of the arm's end-effector is scripted and only D'Hand controls are learned to reduce training time. For the insertion task, the fingers are frozen since it is largely involving accurate motion of the arm.} As in the previous domain, our policy directly controls each joint of the hand and the position of the end-effector. The system is set up to allow for extended uninterrupted operation, allowing for over 30 hours of training without any human intervention. 

\section{Experimental Evaluation}

\begin{figure*}[!ht]
     \centering
     \begin{subfigure}[b]{0.4\textwidth}
         \centering
         \includegraphics[width=.99\textwidth]{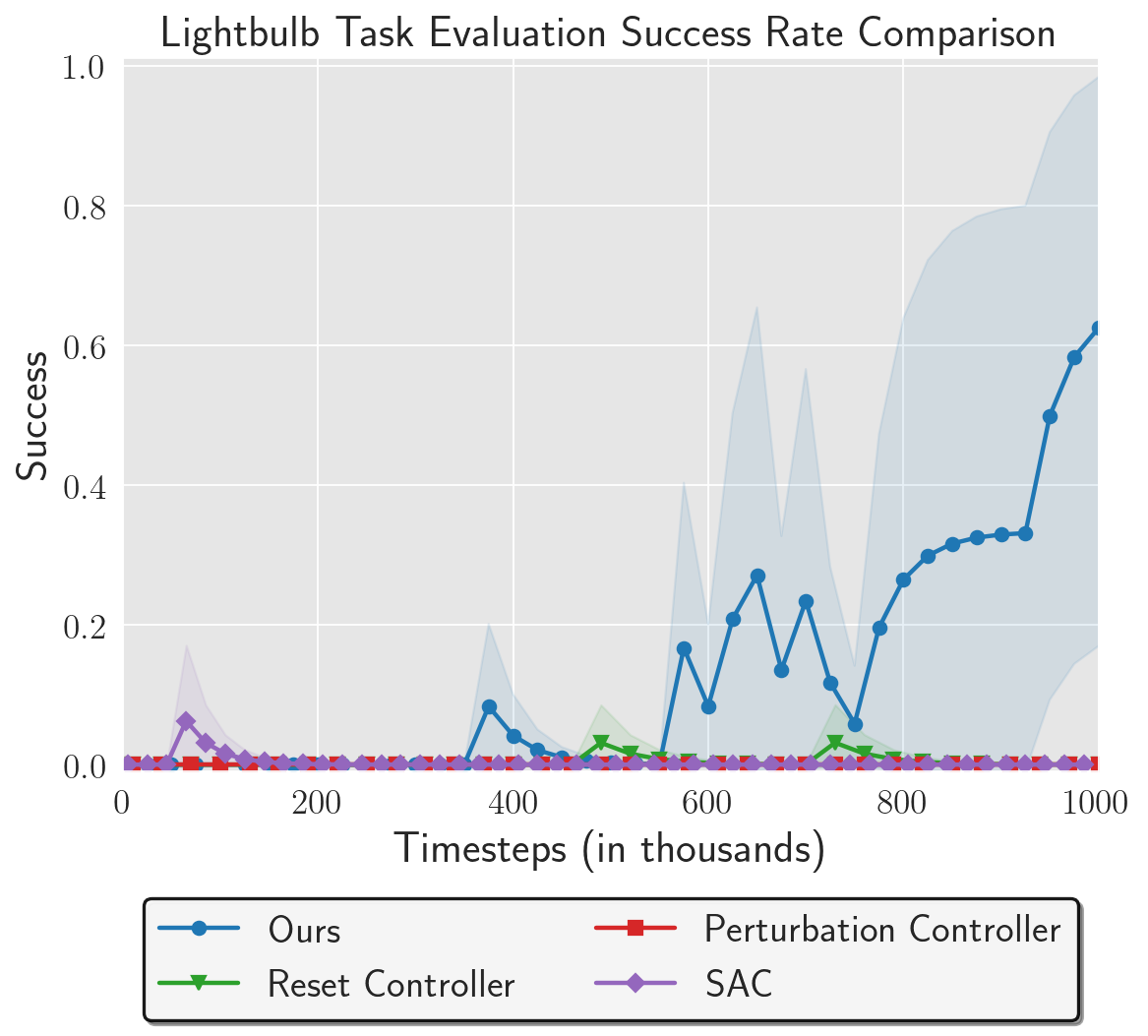}
     \end{subfigure}
    \begin{subfigure}[b]{0.4\textwidth}
         \centering
         \includegraphics[width=.99\textwidth]{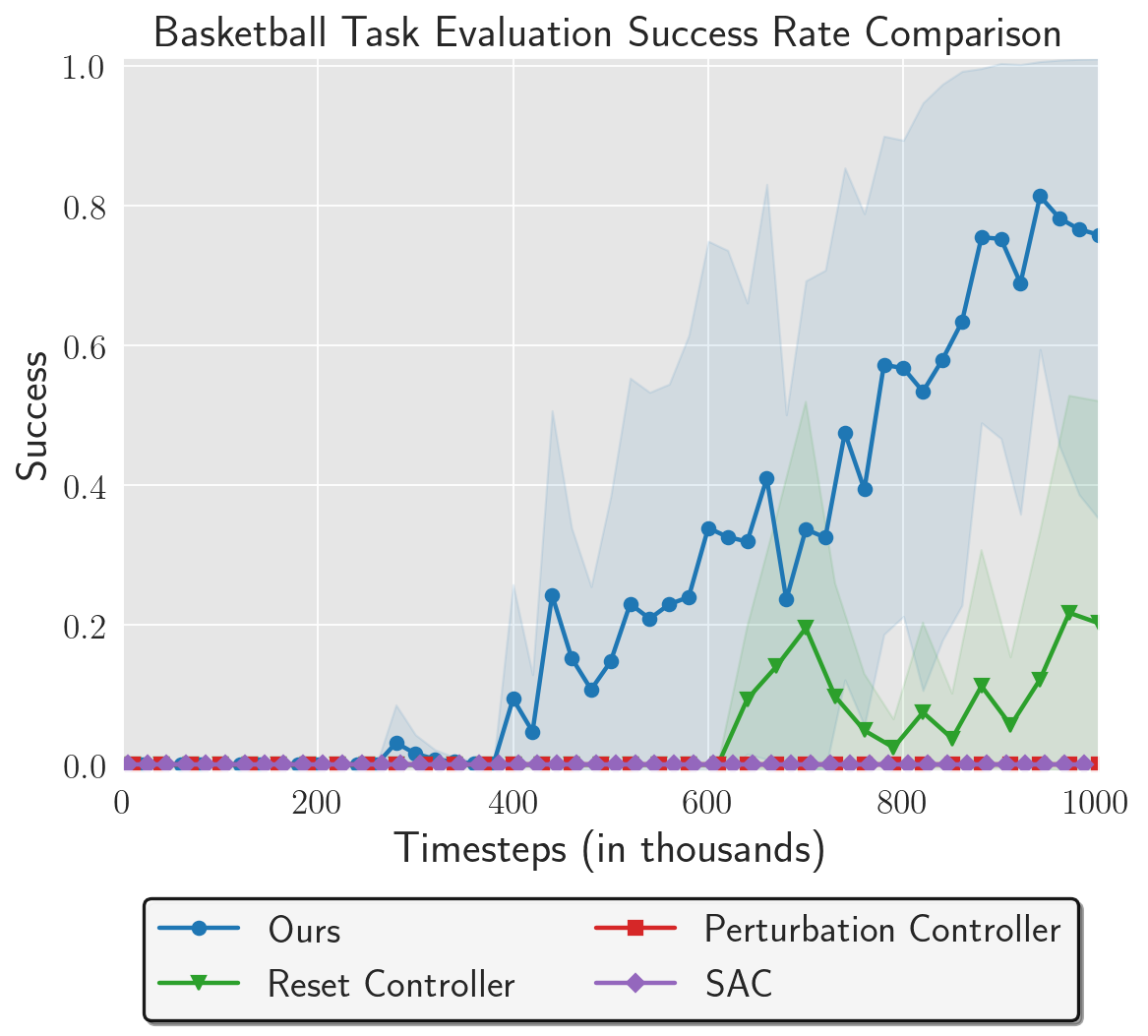}
     \end{subfigure}
     \caption{Comparison of \methodname{} with baseline methods in simulation when run without resets. In comparison to the prior reset-free RL methods, \methodname{} is able to learn the tasks more quickly and with higher average success rates, even in cases where none of the prior methods can master the full task set. \methodname{} is able to solve all of the tasks without requiring any explicit resets.
     }
     \label{fig:results-baselines}
\end{figure*}

We focus our experiment on the following questions:
\begin{enumerate}
    \item Are existing off-policy RL algorithms effective when deployed under reset-free settings to solve dexterous manipulation problems?
    \item Does simultaneously learning a collection of tasks under the proposed multi-task formulation with \methodname{} alleviate the need for resets when solving dexterous manipulation tasks?
    \item Does learning multiple tasks simultaneously allow for reset-free learning of more complex tasks than previous reset free algorithms? 
    \item Does \methodname{} enable real-world reinforcement learning without resets or human interventions?
\end{enumerate}

 \begin{figure}[!h]
 \centering
    \includegraphics[width=0.9\linewidth]{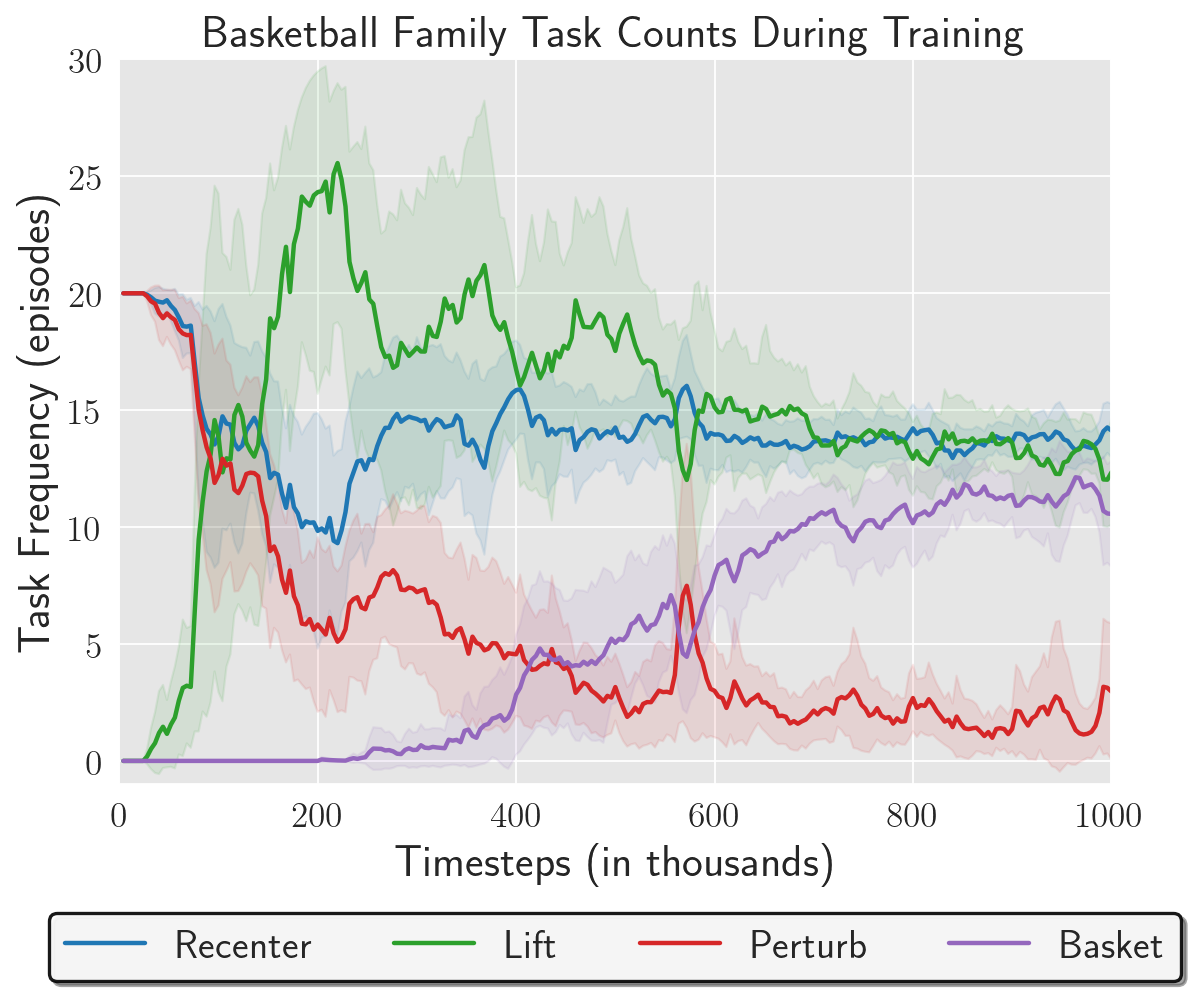}
    \caption{Visualization of task frequency in the basketball task Family. While initially recentering and pickup are common, as these get better they are able to provide resets for other tasks.}
    \label{fig:viz-hand}
 \end{figure}

\subsection{Baselines and Prior Methods}
\label{sec:comparisons}

We compare \methodname{} (Section~\ref{sec:method}) to three prior baseline algorithms. Our first comparison is to a state-of-the-art off-policy RL algorithm, soft actor-critic~\cite{haarnoja2018soft} (labeled as \textit{SAC}). The actor is executed continuously and reset-free in the environment, and the experienced data is stored in a replay pool. This algorithm is representative of efficient off-policy RL algorithms. We next compare to a version of a reset controller~\cite{eysenbach2017leave} (labeled as \textit{Reset Controller}), which trains a forward controller to perform the task and a reset controller to reset the state back to the initial state. Lastly, we compare with the perturbation controller technique~\cite{r3l} introduced in prior work, which alternates between learning and executing a forward task-directed policy and a perturbation controller trained purely with novelty bonuses~\cite{RND} (labeled as \textit{Perturbation Controller}). For all the experiments we used the same RL algorithm, soft actor-critic~\cite{haarnoja2018soft}, with default hyperparameters. To evaluate a task, we roll out its final policy starting from states randomly sampled from the distribution induced by all the tasks that can transition to the task under evaluation, and report performance in terms of their success in solving the task. Additional details, videos of all tasks and hyperparameters can be found at [\siteurl]. 

\begin{figure*}[!ht]
  \centering
  \includegraphics[width=1\textwidth]{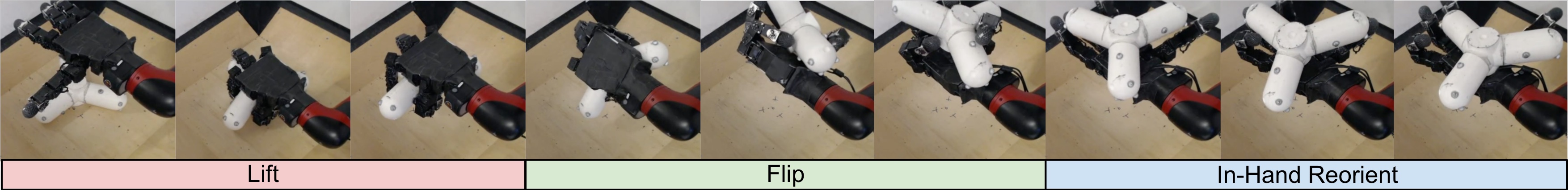}
  \caption{Film strip illustrating partial training trajectory of hardware system for in-hand manipulation of the valve object. This shows various behaviors encountered during the training - picking up the object, flipping it over in the hand and then in-hand manipulation to get it to a particular orientation. As seen here, MTRF is able to successfully learn how to perform in-hand manipulation without any human intervention.}
  \label{fig:example_trajectory_hardware_valve}
\end{figure*}

\begin{figure*}[!ht]
  \centering
  \includegraphics[width=1\textwidth]{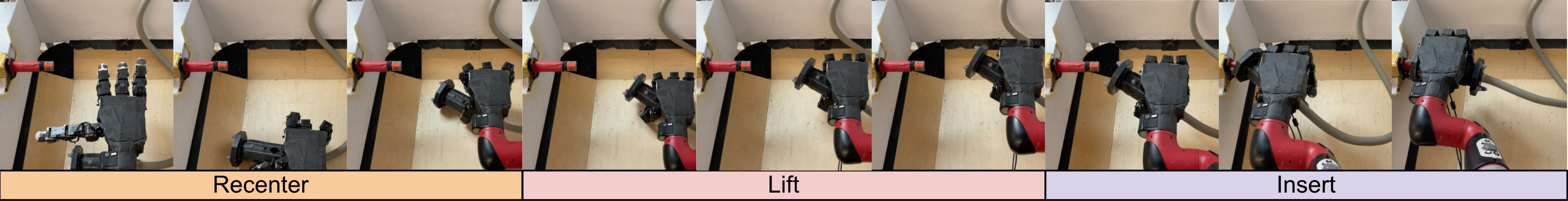}
  \caption{Film strip illustrating partial training trajectory of hardware system for pipe insertion. This shows various behaviors encountered during the training - repositioning the object, picking it up, and then inserting it into the wall attachment. As seen here, MTRF is able to successfully learn how to do pipe insertion without any human intervention.}
  \label{fig:example_trajectory_hardware_pipe}
\end{figure*}

\subsection{Reset-Free Learning Comparisons in Simulation}

We present results for reset-free learning, using our algorithm and prior methods, in Fig~\ref{fig:results-baselines}, corresponding to each of the tasks in simulation in Section~\ref{sec:setup}. We see that \methodname{} is able to successfully learn all of the tasks jointly, as evidenced by Fig~\ref{fig:results-baselines}. We measure evaluation performance after training by loading saved policies and running the policy corresponding to the ``forward'' task for each of the task families (i.e. lightbulb insertion and basketball dunking). This indicates that we can solve all the tasks, and as a result can learn reset free more effectively than prior algorithms.

In comparison, we see that the prior algorithms for off-policy RL -- the reset controller~\cite{eysenbach2017leave} and perturbation controller~\cite{r3l} -- are not able to learn the more complex of the tasks as effectively as our method. While these methods are able to make some amount of progress on tasks that are constructed to be very easy such as the pincer task family shown in Appendix B, we see that they struggle to scale well to the more challenging tasks (Fig~\ref{fig:results-baselines}). Only \methodname{} is able to learn these tasks, while the other methods never actually reach the later tasks in the task graph.

To understand how the tasks are being sequenced during the learning process, we show task transitions experienced during training for the basketball task in Fig~\ref{fig:viz-hand}. We observe that early in training the transitions are mostly between the recenter and perturbation tasks. As \methodname{} improves, the transitions add in pickup and then basketball dunking, cycling between re-centering, pickup and basketball placement in the hoop.

\subsection{Learning Real-World Dexterous Manipulation Skills}

Next, we evaluate the performance of MTRF on the real-world robotic system described in Section~\ref{sec:setup}, studying the dexterous manipulation tasks described in Section~\ref{sec:hardware-setup}. 

\paragraph{In-Hand Manipulation} Let us start by considering the in-hand manipulation tasks shown in Fig~\ref{fig:inhand-taskgraph}. This task is challenging because it requires delicate handling of finger-object contacts during training, the object is easy to drop during the flip-up and in-hand manipulation, and the reorientation requires a coordinated finger gait to rotate the object. In fact, most prior work that aims to learn similar in-hand manipulation behaviors either utilizes simulation or employs a hand-designed reset procedure, potentially involving human interventions~\cite{pddm, zhu2019dexterous, kumaroptimal}. To the best of our knowledge, our work is the first to show a real-world robotic system learning such a task entirely in the real world and without any manually provided or hand-designed reset mechanism. We visualize a sequential execution of the tasks (after training) in Fig~\ref{fig:example_trajectory_hardware_valve}, and encourage the reader to watch a video of this task, as well as the training process, on the project website: \mbox{[\siteurl]}. Over the course of training, the robot must first learn to recenter the object on the table, then learn to pick it up (which requires learning an appropriate grasp and delicate control of the fingers to maintain grip), then learn to flip up the object so that it rests in the palm, and finally learn to perform the orientation. Dropping the object at any point in this process requires going back to the beginning of the sequence, and initially most of the training time is spent on re-centering, which provides resets for the pickup. The entire training process takes about 60 hours of real time, learning all of the tasks simultaneously. Although this time requirement is considerable, it is entirely autonomous, making this approach scalable even without any simulation or manual instrumentation. The user only needs to position the objects for training, and switch on the robot.

\begin{figure}[!h]
     \centering
        \includegraphics[width=0.49\linewidth]{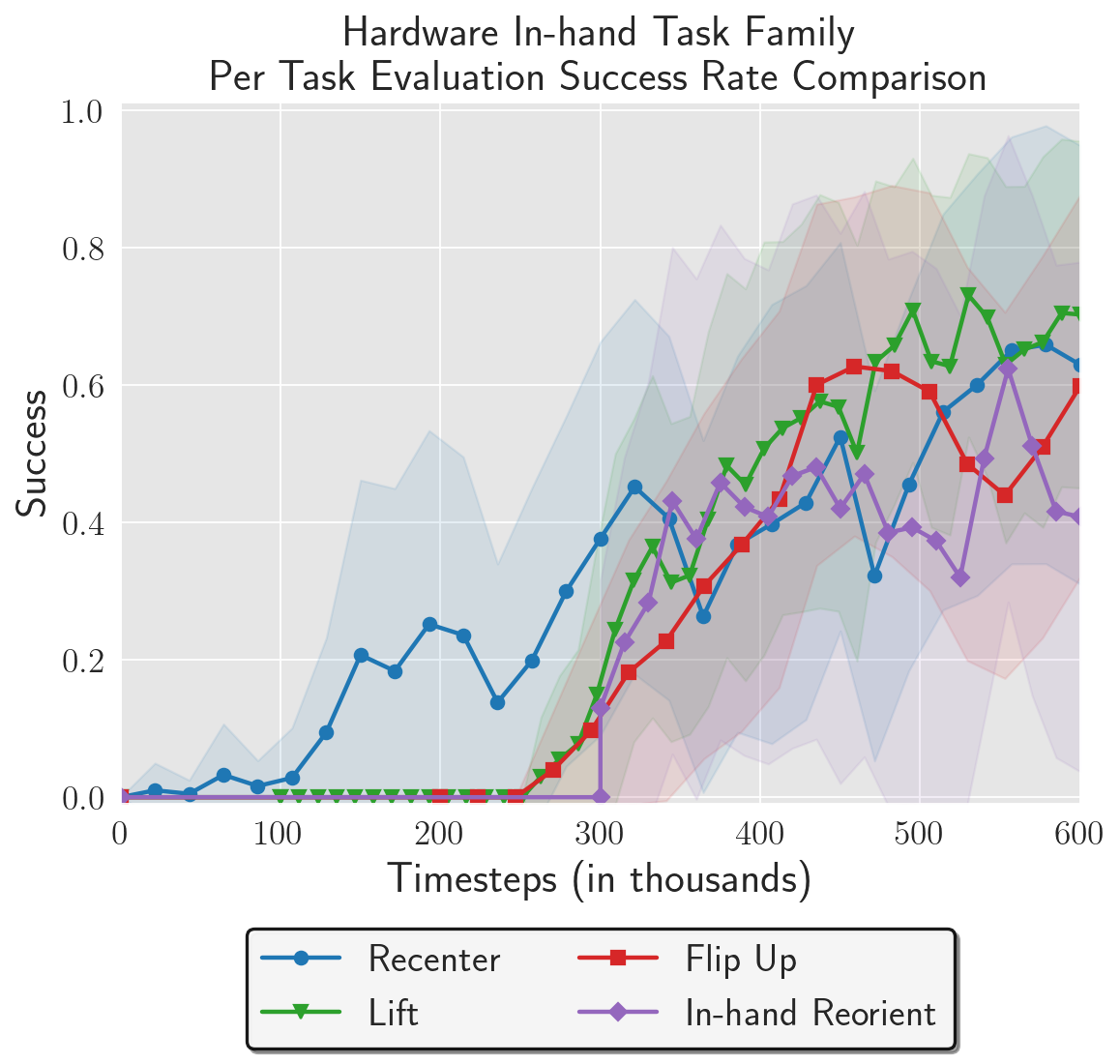}
        \includegraphics[width=0.49\linewidth]{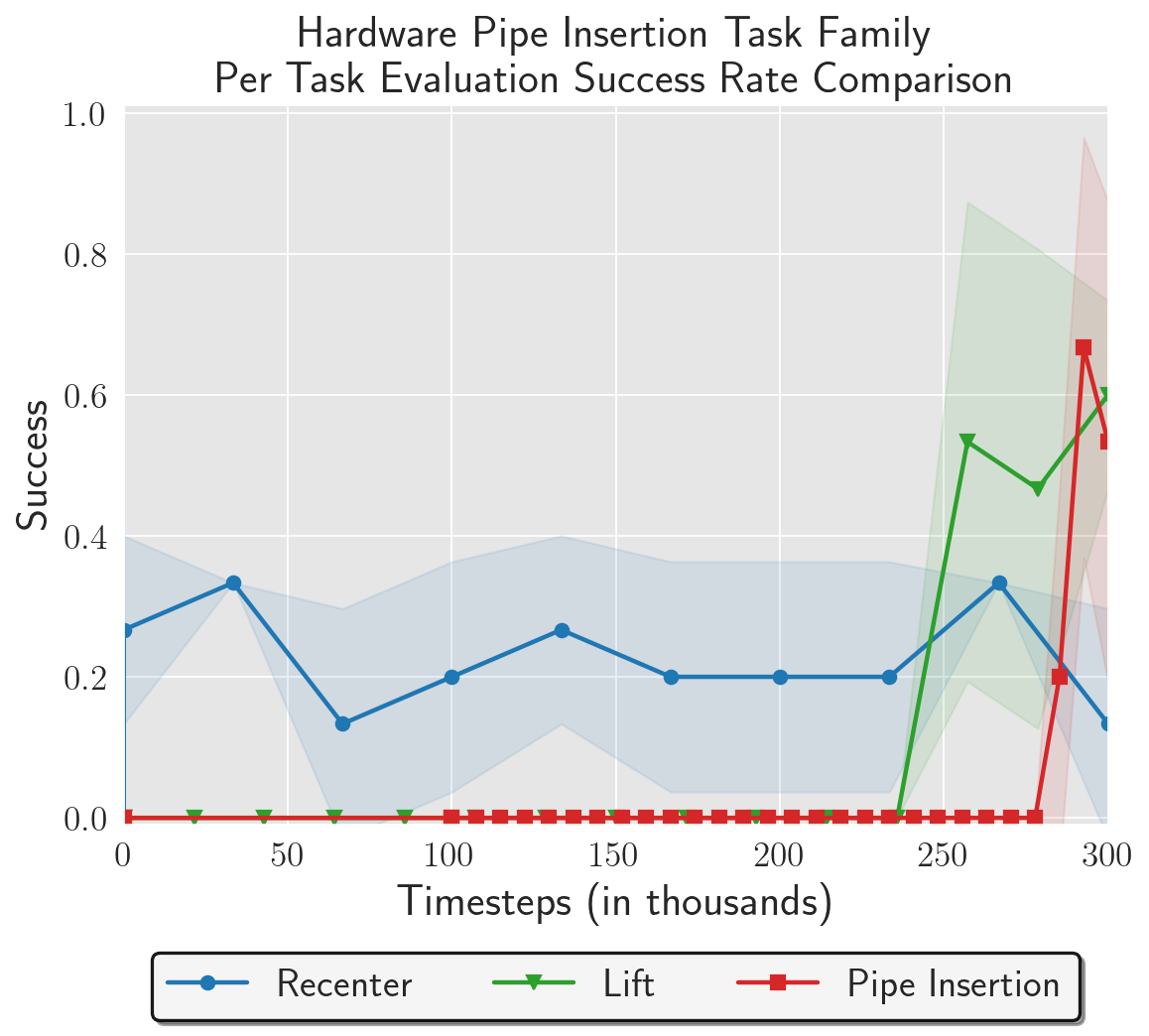}
        \caption{Success rate of various tasks on dexterous manipulation task families on hardware. \textbf{Left: In-hand manipulation} We can see that all of the tasks are able to successfully learn with more    than $70\%$ success rate. \textbf{Right: Pipe insertion} We can see that the final pipe insertion task is able to successfully learn with more than $60\%$ success rate. }
        \label{fig:hardware_success}
\end{figure}

For a quantitative evaluation, we plot the success rate of sub-tasks including re-centering, lifting, flipping over, and in-hand reorientation. For lifting and flipping over, success is defined as lifting the object to a particular height above the table, and for reorient success is defined by the difference between the current object orientation and the target orientation of the object. As shown in Fig~\ref{fig:hardware_success}, \methodname{} is able to autonomously learn all tasks in the task graph in 60 hours, and achieves an 70\% success rate for the in-hand reorient task. This experiment illustrates how \methodname{} can enable a complex real-world robotic system to learn an in-hand manipulation behavior while at the same time autonomously retrying the task during a lengthy unattended training run, without any simulation, special instrumentation, or manual interventions. This experiment suggests that, when \methodname{} is provided with an appropriate set of tasks, learning of complex manipulation skills can be carried out entirely autonomously in the real world, even for highly complex robotic manipulators such as multi-fingered hands.

\paragraph{Pipe Insertion} We also considered the second task variant which involves manipulating a cylindrical pipe to insert it into a hose attachment on the wall as shown in Fig~\ref{fig:pipe-taskgraph}. This task is challenging because it requires accurate grasping and repositioning of the object during training in order to accurately insert it into the hose attachment, requiring coordination of both the arm and the hand. Training this task without resets requires a combination of repositioning, lifting, insertion and removal to continually keep training and improving. We visualize a sequential execution of the tasks (after training) in Fig~\ref{fig:example_trajectory_hardware_pipe}, and encourage the reader to watch a video of this task, as well as the training process, on the project website: \mbox{[\siteurl]}. Over the course of training, the robot must first learn to recenter the object on the table, then learn to pick it up (which requires learning an appropriate grasp), then learn to actually move the arm accurately to insert the pipe to the attachment, and finally learn to remove it to continue training and practicing. Initially most of the training time is spent on re-centering, which provides resets for the pickup, which then provides resets for the insertion and removal. The entire training process takes about 25 hours of real time, learning all of the tasks simultaneously.

\section{Discussion}
In this work, we introduced a technique for learning dexterous manipulation behaviors reset-free, without the need for any human intervention during training. This was done by leveraging the multi-task RL setting for reset-free learning. When learning multiple tasks simultaneously, different tasks serve to reset each other and assist in uninterrupted learning. This algorithm allows a dexterous manipulation system to learn manipulation behaviors uninterrupted, and also learn behavior that allows it to continue practicing. Our experiments show that this approach can enable a real-world hand-arm robotic system to learn an in-hand reorientation task, including pickup and repositioning, in about 60 hours of training, as well as a pipe insertion task in around 25 hours of training without human intervention or special instrumentation.

\section{Acknowledgements} 
The authors would like to thank Anusha Nagabandi, Greg Kahn, Archit Sharma, Aviral Kumar, Benjamin Eysenbach, Karol Hausman, Corey Lynch for helpful discussions and comments.This research was supported by the Office of Naval Research, the National Science Foundation, and Berkeley DeepDrive.

\bibliographystyle{IEEEtran}
\bibliography{references}

\begin{thebibliography}{10}
\providecommand{\url}[1]{#1}
\csname url@samestyle\endcsname
\providecommand{\newblock}{\relax}
\providecommand{\bibinfo}[2]{#2}
\providecommand{\BIBentrySTDinterwordspacing}{\spaceskip=0pt\relax}
\providecommand{\BIBentryALTinterwordstretchfactor}{4}
\providecommand{\BIBentryALTinterwordspacing}{\spaceskip=\fontdimen2\font plus
\BIBentryALTinterwordstretchfactor\fontdimen3\font minus
  \fontdimen4\font\relax}
\providecommand{\BIBforeignlanguage}[2]{{%
\expandafter\ifx\csname l@#1\endcsname\relax
\typeout{** WARNING: IEEEtran.bst: No hyphenation pattern has been}%
\typeout{** loaded for the language `#1'. Using the pattern for}%
\typeout{** the default language instead.}%
\else
\language=\csname l@#1\endcsname
\fi
#2}}
\providecommand{\BIBdecl}{\relax}
\BIBdecl

\bibitem{DAPG}
\BIBentryALTinterwordspacing
A.~Rajeswaran, V.~Kumar, A.~Gupta, J.~Schulman, E.~Todorov, and S.~Levine,
  ``Learning complex dexterous manipulation with deep reinforcement learning
  and demonstrations,'' \emph{CoRR}, vol. abs/1709.10087, 2017. [Online].
  Available: \url{http://arxiv.org/abs/1709.10087}
\BIBentrySTDinterwordspacing

\bibitem{2018-TOG-deepMimic}
\BIBentryALTinterwordspacing
X.~B. Peng, P.~Abbeel, S.~Levine, and M.~van~de Panne, ``Deepmimic:
  Example-guided deep reinforcement learning of physics-based character
  skills,'' \emph{ACM Trans. Graph.}, vol.~37, no.~4, pp. 143:1--143:14, Jul.
  2018. [Online]. Available: \url{http://doi.acm.org/10.1145/3197517.3201311}
\BIBentrySTDinterwordspacing

\bibitem{levine2015learning}
S.~Levine, N.~Wagener, and P.~Abbeel, ``Learning contact-rich manipulation
  skills with guided policy search,'' in \emph{Robotics and Automation (ICRA),
  2015 IEEE International Conference on}.\hskip 1em plus 0.5em minus
  0.4em\relax IEEE, 2015.

\bibitem{pinto2016supersizing}
L.~Pinto and A.~Gupta, ``Supersizing self-supervision: Learning to grasp from
  50k tries and 700 robot hours,'' in \emph{2016 IEEE international conference
  on robotics and automation (ICRA)}.\hskip 1em plus 0.5em minus 0.4em\relax
  IEEE, 2016, pp. 3406--3413.

\bibitem{andrychowicz2020learning}
O.~M. Andrychowicz, B.~Baker, M.~Chociej, R.~Jozefowicz, B.~McGrew,
  J.~Pachocki, A.~Petron, M.~Plappert, G.~Powell, A.~Ray \emph{et~al.},
  ``Learning dexterous in-hand manipulation,'' \emph{The International Journal
  of Robotics Research}, vol.~39, no.~1, pp. 3--20, 2020.

\bibitem{qtopt}
D.~Kalashnikov, A.~Irpan, P.~Pastor, J.~Ibarz, A.~Herzog, E.~Jang, D.~Quillen,
  E.~Holly, M.~Kalakrishnan, V.~Vanhoucke \emph{et~al.}, ``Qt-opt: Scalable
  deep reinforcement learning for vision-based robotic manipulation,''
  \emph{arXiv preprint arXiv:1806.10293}, 2018.

\bibitem{pddm}
A.~Nagabandi, K.~Konolige, S.~Levine, and V.~Kumar, ``Deep dynamics models for
  learning dexterous manipulation,'' in \emph{Conference on Robot Learning},
  2020, pp. 1101--1112.

\bibitem{r3l}
H.~Zhu, J.~Yu, A.~Gupta, D.~Shah, K.~Hartikainen, A.~Singh, V.~Kumar, and
  S.~Levine, ``The ingredients of real-world robotic reinforcement learning,''
  \emph{arXiv preprint arXiv:2004.12570}, 2020.

\bibitem{ploeger2020high}
K.~Ploeger, M.~Lutter, and J.~Peters, ``High acceleration reinforcement
  learning for real-world juggling with binary rewards,'' 2020.

\bibitem{kumaroptimal}
\BIBentryALTinterwordspacing
V.~Kumar, E.~Todorov, and S.~Levine, ``Optimal control with learned local
  models: Application to dexterous manipulation,'' in \emph{2016 {IEEE}
  International Conference on Robotics and Automation, {ICRA} 2016, Stockholm,
  Sweden, May 16-21, 2016}, D.~Kragic, A.~Bicchi, and A.~D. Luca, Eds.\hskip
  1em plus 0.5em minus 0.4em\relax {IEEE}, 2016, pp. 378--383. [Online].
  Available: \url{https://doi.org/10.1109/ICRA.2016.7487156}
\BIBentrySTDinterwordspacing

\bibitem{dppt}
\BIBentryALTinterwordspacing
A.~Ghadirzadeh, A.~Maki, D.~Kragic, and M.~Bj{\"{o}}rkman, ``Deep predictive
  policy training using reinforcement learning,'' \emph{CoRR}, vol.
  abs/1703.00727, 2017. [Online]. Available:
  \url{http://arxiv.org/abs/1703.00727}
\BIBentrySTDinterwordspacing

\bibitem{zhu2019dexterous}
H.~Zhu, A.~Gupta, A.~Rajeswaran, S.~Levine, and V.~Kumar, ``Dexterous
  manipulation with deep reinforcement learning: Efficient, general, and
  low-cost,'' in \emph{2019 International Conference on Robotics and Automation
  (ICRA)}.\hskip 1em plus 0.5em minus 0.4em\relax IEEE, 2019, pp. 3651--3657.

\bibitem{pi2gps}
\BIBentryALTinterwordspacing
Y.~Chebotar, M.~Kalakrishnan, A.~Yahya, A.~Li, S.~Schaal, and S.~Levine, ``Path
  integral guided policy search,'' \emph{CoRR}, vol. abs/1610.00529, 2016.
  [Online]. Available: \url{http://arxiv.org/abs/1610.00529}
\BIBentrySTDinterwordspacing

\bibitem{eysenbach2017leave}
B.~Eysenbach, S.~Gu, J.~Ibarz, and S.~Levine, ``Leave no trace: Learning to
  reset for safe and autonomous reinforcement learning,'' \emph{arXiv preprint
  arXiv:1711.06782}, 2017.

\bibitem{ahn2020robel}
M.~Ahn, H.~Zhu, K.~Hartikainen, H.~Ponte, A.~Gupta, S.~Levine, and V.~Kumar,
  ``Robel: Robotics benchmarks for learning with low-cost robots,'' in
  \emph{Conference on Robot Learning}.\hskip 1em plus 0.5em minus 0.4em\relax
  PMLR, 2020, pp. 1300--1313.

\bibitem{pulkit}
\BIBentryALTinterwordspacing
P.~Agrawal, A.~Nair, P.~Abbeel, J.~Malik, and S.~Levine, ``Learning to poke by
  poking: Experiential learning of intuitive physics,'' \emph{CoRR}, vol.
  abs/1606.07419, 2016. [Online]. Available:
  \url{http://arxiv.org/abs/1606.07419}
\BIBentrySTDinterwordspacing

\bibitem{asymmetricselfplay}
S.~Sukhbaatar, Z.~Lin, I.~Kostrikov, G.~Synnaeve, A.~Szlam, and R.~Fergus,
  ``Intrinsic motivation and automatic curricula via asymmetric self-play,''
  \emph{arXiv preprint arXiv:1703.05407}, 2017.

\bibitem{richter17visual}
\BIBentryALTinterwordspacing
C.~Richter and N.~Roy, ``Safe visual navigation via deep learning and novelty
  detection,'' in \emph{Robotics: Science and Systems XIII, Massachusetts
  Institute of Technology, Cambridge, Massachusetts, USA, July 12-16, 2017},
  N.~M. Amato, S.~S. Srinivasa, N.~Ayanian, and S.~Kuindersma, Eds., 2017.
  [Online]. Available: \url{http://www.roboticsproceedings.org/rss13/p64.html}
\BIBentrySTDinterwordspacing

\bibitem{distral}
\BIBentryALTinterwordspacing
Y.~W. Teh, V.~Bapst, W.~M. Czarnecki, J.~Quan, J.~Kirkpatrick, R.~Hadsell,
  N.~Heess, and R.~Pascanu, ``Distral: Robust multitask reinforcement
  learning,'' in \emph{Advances in Neural Information Processing Systems 30:
  Annual Conference on Neural Information Processing Systems 2017, 4-9 December
  2017, Long Beach, CA, {USA}}, I.~Guyon, U.~von Luxburg, S.~Bengio, H.~M.
  Wallach, R.~Fergus, S.~V.~N. Vishwanathan, and R.~Garnett, Eds., 2017, pp.
  4496--4506. [Online]. Available:
  \url{http://papers.nips.cc/paper/7036-distral-robust-multitask-reinforcement-learning}
\BIBentrySTDinterwordspacing

\bibitem{policydistillation}
\BIBentryALTinterwordspacing
A.~A. Rusu, S.~G. Colmenarejo, {\c{C}}.~G{\"{u}}l{\c{c}}ehre, G.~Desjardins,
  J.~Kirkpatrick, R.~Pascanu, V.~Mnih, K.~Kavukcuoglu, and R.~Hadsell, ``Policy
  distillation,'' in \emph{4th International Conference on Learning
  Representations, {ICLR} 2016, San Juan, Puerto Rico, May 2-4, 2016,
  Conference Track Proceedings}, Y.~Bengio and Y.~LeCun, Eds., 2016. [Online].
  Available: \url{http://arxiv.org/abs/1511.06295}
\BIBentrySTDinterwordspacing

\bibitem{actor-mimic}
\BIBentryALTinterwordspacing
E.~Parisotto, L.~J. Ba, and R.~Salakhutdinov, ``Actor-mimic: Deep multitask and
  transfer reinforcement learning,'' in \emph{4th International Conference on
  Learning Representations, {ICLR} 2016, San Juan, Puerto Rico, May 2-4, 2016,
  Conference Track Proceedings}, Y.~Bengio and Y.~LeCun, Eds., 2016. [Online].
  Available: \url{http://arxiv.org/abs/1511.06342}
\BIBentrySTDinterwordspacing

\bibitem{pcgrad}
\BIBentryALTinterwordspacing
T.~Yu, S.~Kumar, A.~Gupta, S.~Levine, K.~Hausman, and C.~Finn, ``Gradient
  surgery for multi-task learning,'' \emph{CoRR}, vol. abs/2001.06782, 2020.
  [Online]. Available: \url{https://arxiv.org/abs/2001.06782}
\BIBentrySTDinterwordspacing

\bibitem{ozansener}
\BIBentryALTinterwordspacing
O.~Sener and V.~Koltun, ``Multi-task learning as multi-objective
  optimization,'' \emph{CoRR}, vol. abs/1810.04650, 2018. [Online]. Available:
  \url{http://arxiv.org/abs/1810.04650}
\BIBentrySTDinterwordspacing

\bibitem{meta-world}
\BIBentryALTinterwordspacing
T.~Yu, D.~Quillen, Z.~He, R.~Julian, K.~Hausman, C.~Finn, and S.~Levine,
  ``Meta-world: {A} benchmark and evaluation for multi-task and meta
  reinforcement learning,'' in \emph{3rd Annual Conference on Robot Learning,
  CoRL 2019, Osaka, Japan, October 30 - November 1, 2019, Proceedings}, ser.
  Proceedings of Machine Learning Research, L.~P. Kaelbling, D.~Kragic, and
  K.~Sugiura, Eds., vol. 100.\hskip 1em plus 0.5em minus 0.4em\relax {PMLR},
  2019, pp. 1094--1100. [Online]. Available:
  \url{http://proceedings.mlr.press/v100/yu20a.html}
\BIBentrySTDinterwordspacing

\bibitem{james2019rlbench}
S.~James, Z.~Ma, D.~R. Arrojo, and A.~J. Davison, ``Rlbench: The robot learning
  benchmark and learning environment,'' 2019.

\bibitem{deepmindheess}
\BIBentryALTinterwordspacing
N.~Heess, D.~TB, S.~Sriram, J.~Lemmon, J.~Merel, G.~Wayne, Y.~Tassa, T.~Erez,
  Z.~Wang, S.~M.~A. Eslami, M.~A. Riedmiller, and D.~Silver, ``Emergence of
  locomotion behaviours in rich environments,'' \emph{CoRR}, vol.
  abs/1707.02286, 2017. [Online]. Available:
  \url{http://arxiv.org/abs/1707.02286}
\BIBentrySTDinterwordspacing

\bibitem{openAIhand}
\BIBentryALTinterwordspacing
OpenAI, M.~Andrychowicz, B.~Baker, M.~Chociej, R.~J{\'{o}}zefowicz, B.~McGrew,
  J.~W. Pachocki, J.~Pachocki, A.~Petron, M.~Plappert, G.~Powell, A.~Ray,
  J.~Schneider, S.~Sidor, J.~Tobin, P.~Welinder, L.~Weng, and W.~Zaremba,
  ``Learning dexterous in-hand manipulation,'' \emph{CoRR}, vol.
  abs/1808.00177, 2018. [Online]. Available:
  \url{http://arxiv.org/abs/1808.00177}
\BIBentrySTDinterwordspacing

\bibitem{ha2020learning}
S.~Ha, P.~Xu, Z.~Tan, S.~Levine, and J.~Tan, ``Learning to walk in the real
  world with minimal human effort,'' 2020.

\bibitem{peng2020learning}
X.~B. Peng, E.~Coumans, T.~Zhang, T.-W. Lee, J.~Tan, and S.~Levine, ``Learning
  agile robotic locomotion skills by imitating animals,'' 2020.

\bibitem{calandra2016gait}
\BIBentryALTinterwordspacing
R.~Calandra, A.~Seyfarth, J.~Peters, and M.~P. Deisenroth, ``Bayesian
  optimization for learning gaits under uncertainty - an experimental
  comparison on a dynamic bipedal walker,'' \emph{Ann. Math. Artif. Intell.},
  vol.~76, no. 1-2, pp. 5--23, 2016. [Online]. Available:
  \url{https://doi.org/10.1007/s10472-015-9463-9}
\BIBentrySTDinterwordspacing

\bibitem{handeye}
\BIBentryALTinterwordspacing
S.~Levine, P.~Pastor, A.~Krizhevsky, and D.~Quillen, ``Learning hand-eye
  coordination for robotic grasping with deep learning and large-scale data
  collection,'' \emph{CoRR}, vol. abs/1603.02199, 2016. [Online]. Available:
  \url{http://arxiv.org/abs/1603.02199}
\BIBentrySTDinterwordspacing

\bibitem{baier07grasping}
\BIBentryALTinterwordspacing
T.~Baier{-}L{\"{o}}wenstein and J.~Zhang, ``Learning to grasp everyday objects
  using reinforcement-learning with automatic value cut-off,'' in \emph{2007
  {IEEE/RSJ} International Conference on Intelligent Robots and Systems,
  October 29 - November 2, 2007, Sheraton Hotel and Marina, San Diego,
  California, {USA}}.\hskip 1em plus 0.5em minus 0.4em\relax {IEEE}, 2007, pp.
  1551--1556. [Online]. Available:
  \url{https://doi.org/10.1109/IROS.2007.4399053}
\BIBentrySTDinterwordspacing

\bibitem{wu2019mat}
B.~Wu, I.~Akinola, J.~Varley, and P.~Allen, ``Mat: Multi-fingered adaptive
  tactile grasping via deep reinforcement learning,'' 2019.

\bibitem{nemec2017door}
\BIBentryALTinterwordspacing
B.~Nemec, L.~Zlajpah, and A.~Ude, ``Door opening by joining reinforcement
  learning and intelligent control,'' in \emph{18th International Conference on
  Advanced Robotics, {ICAR} 2017, Hong Kong, China, July 10-12, 2017}.\hskip
  1em plus 0.5em minus 0.4em\relax {IEEE}, 2017, pp. 222--228. [Online].
  Available: \url{https://doi.org/10.1109/ICAR.2017.8023522}
\BIBentrySTDinterwordspacing

\bibitem{doorgym}
\BIBentryALTinterwordspacing
Y.~Urakami, A.~Hodgkinson, C.~Carlin, R.~Leu, L.~Rigazio, and P.~Abbeel,
  ``Doorgym: {A} scalable door opening environment and baseline agent,''
  \emph{CoRR}, vol. abs/1908.01887, 2019. [Online]. Available:
  \url{http://arxiv.org/abs/1908.01887}
\BIBentrySTDinterwordspacing

\bibitem{ddpgshane}
\BIBentryALTinterwordspacing
S.~Gu, E.~Holly, T.~P. Lillicrap, and S.~Levine, ``Deep reinforcement learning
  for robotic manipulation,'' \emph{CoRR}, vol. abs/1610.00633, 2016. [Online].
  Available: \url{http://arxiv.org/abs/1610.00633}
\BIBentrySTDinterwordspacing

\bibitem{RIG}
\BIBentryALTinterwordspacing
A.~Nair, V.~Pong, M.~Dalal, S.~Bahl, S.~Lin, and S.~Levine, ``Visual
  reinforcement learning with imagined goals,'' in \emph{Advances in Neural
  Information Processing Systems 31: Annual Conference on Neural Information
  Processing Systems 2018, NeurIPS 2018, 3-8 December 2018, Montr{\'{e}}al,
  Canada}, S.~Bengio, H.~M. Wallach, H.~Larochelle, K.~Grauman,
  N.~Cesa{-}Bianchi, and R.~Garnett, Eds., 2018, pp. 9209--9220. [Online].
  Available:
  \url{http://papers.nips.cc/paper/8132-visual-reinforcement-learning-with-imagined-goals}
\BIBentrySTDinterwordspacing

\bibitem{vanHoof2015learning}
H.~van Hoof, T.~Hermans, G.~Neumann, and J.~Peters, ``Learning robot in-hand
  manipulation with tactile features,'' in \emph{Humanoid Robots
  (Humanoids)}.\hskip 1em plus 0.5em minus 0.4em\relax IEEE, 2015.

\bibitem{okamura2000overview}
A.~M. Okamura, N.~Smaby, and M.~R. Cutkosky, ``An overview of dexterous
  manipulation,'' in \emph{Robotics and Automation, 2000. Proceedings. ICRA'00.
  IEEE International Conference on}, vol.~1.\hskip 1em plus 0.5em minus
  0.4em\relax IEEE, 2000, pp. 255--262.

\bibitem{furukawa2006dynamic}
N.~Furukawa, A.~Namiki, S.~Taku, and M.~Ishikawa, ``Dynamic regrasping using a
  high-speed multifingered hand and a high-speed vision system,'' in
  \emph{Proceedings 2006 IEEE International Conference on Robotics and
  Automation, 2006. ICRA 2006.}\hskip 1em plus 0.5em minus 0.4em\relax IEEE,
  2006, pp. 181--187.

\bibitem{bai2014dexterous}
Y.~Bai and C.~K. Liu, ``Dexterous manipulation using both palm and fingers,''
  in \emph{ICRA 2014}.\hskip 1em plus 0.5em minus 0.4em\relax IEEE, 2014.

\bibitem{mordatch2012contact}
I.~Mordatch, Z.~Popovi{\'c}, and E.~Todorov, ``Contact-invariant optimization
  for hand manipulation,'' in \emph{Proceedings of the ACM
  SIGGRAPH/Eurographics symposium on computer animation}.\hskip 1em plus 0.5em
  minus 0.4em\relax Eurographics Association, 2012.

\bibitem{mpcvikash}
\BIBentryALTinterwordspacing
V.~Kumar, Y.~Tassa, T.~Erez, and E.~Todorov, ``Real-time behaviour synthesis
  for dynamic hand-manipulation,'' in \emph{2014 {IEEE} International
  Conference on Robotics and Automation, {ICRA} 2014, Hong Kong, China, May 31
  - June 7, 2014}, 2014, pp. 6808--6815. [Online]. Available:
  \url{https://doi.org/10.1109/ICRA.2014.6907864}
\BIBentrySTDinterwordspacing

\bibitem{yamane2004synthesizing}
K.~Yamane, J.~J. Kuffner, and J.~K. Hodgins, ``Synthesizing animations of human
  manipulation tasks,'' in \emph{ACM SIGGRAPH 2004 Papers}.\hskip 1em plus
  0.5em minus 0.4em\relax CRC press, 2004, pp. 532--539.

\bibitem{mandikal2020dexterous}
P.~Mandikal and K.~Grauman, ``Dexterous robotic grasping with object-centric
  visual affordances,'' 2020.

\bibitem{jain2019learning}
V.~Kumar, A.~Gupta, E.~Todorov, and S.~Levine, ``Learning dexterous
  manipulation policies from experience and imitation,'' \emph{arXiv preprint
  arXiv:1611.05095}, 2016.

\bibitem{dexterousopenai}
\BIBentryALTinterwordspacing
OpenAI, ``Learning dexterous in-hand manipulation,'' \emph{CoRR}, vol.
  abs/1808.00177, 2018. [Online]. Available:
  \url{http://arxiv.org/abs/1808.00177}
\BIBentrySTDinterwordspacing

\bibitem{vanhoof}
\BIBentryALTinterwordspacing
H.~van Hoof, T.~Hermans, G.~Neumann, and J.~Peters, ``Learning robot in-hand
  manipulation with tactile features,'' in \emph{15th {IEEE-RAS} International
  Conference on Humanoid Robots, Humanoids 2015, Seoul, South Korea, November
  3-5, 2015}.\hskip 1em plus 0.5em minus 0.4em\relax {IEEE}, 2015, pp.
  121--127. [Online]. Available:
  \url{https://doi.org/10.1109/HUMANOIDS.2015.7363524}
\BIBentrySTDinterwordspacing

\bibitem{softhand}
\BIBentryALTinterwordspacing
A.~Gupta, C.~Eppner, S.~Levine, and P.~Abbeel, ``Learning dexterous
  manipulation for a soft robotic hand from human demonstrations,'' in
  \emph{2016 {IEEE/RSJ} International Conference on Intelligent Robots and
  Systems, {IROS} 2016, Daejeon, South Korea, October 9-14, 2016}, 2016, pp.
  3786--3793. [Online]. Available:
  \url{https://doi.org/10.1109/IROS.2016.7759557}
\BIBentrySTDinterwordspacing

\bibitem{choisofthand}
C.~Choi, W.~Schwarting, J.~DelPreto, and D.~Rus, ``Learning object grasping for
  soft robot hands,'' \emph{IEEE Robotics and Automation Letters}, vol.~3,
  no.~3, pp. 2370--2377, 2018.

\bibitem{kumar2016learning}
V.~Kumar, A.~Gupta, E.~Todorov, and S.~Levine, ``Learning dexterous
  manipulation policies from experience and imitation,'' \emph{arXiv preprint
  arXiv:1611.05095}, 2016.

\bibitem{rfgps}
\BIBentryALTinterwordspacing
W.~Montgomery, A.~Ajay, C.~Finn, P.~Abbeel, and S.~Levine, ``Reset-free guided
  policy search: Efficient deep reinforcement learning with stochastic initial
  states,'' \emph{CoRR}, vol. abs/1610.01112, 2016. [Online]. Available:
  \url{http://arxiv.org/abs/1610.01112}
\BIBentrySTDinterwordspacing

\bibitem{han2015learning}
W.~Han, S.~Levine, and P.~Abbeel, ``Learning compound multi-step controllers
  under unknown dynamics,'' in \emph{2015 IEEE/RSJ International Conference on
  Intelligent Robots and Systems (IROS)}.\hskip 1em plus 0.5em minus
  0.4em\relax IEEE, 2015, pp. 6435--6442.

\bibitem{smith2019avid}
L.~Smith, N.~Dhawan, M.~Zhang, P.~Abbeel, and S.~Levine, ``Avid: Learning
  multi-stage tasks via pixel-level translation of human videos,'' \emph{arXiv
  preprint arXiv:1912.04443}, 2019.

\bibitem{rudermtl}
\BIBentryALTinterwordspacing
S.~Ruder, ``An overview of multi-task learning in deep neural networks,''
  \emph{CoRR}, vol. abs/1706.05098, 2017. [Online]. Available:
  \url{http://arxiv.org/abs/1706.05098}
\BIBentrySTDinterwordspacing

\bibitem{yang2020mtrl}
\BIBentryALTinterwordspacing
R.~Yang, H.~Xu, Y.~Wu, and X.~Wang, ``Multi-task reinforcement learning with
  soft modularization,'' \emph{CoRR}, vol. abs/2003.13661, 2020. [Online].
  Available: \url{https://arxiv.org/abs/2003.13661}
\BIBentrySTDinterwordspacing

\bibitem{sacx}
\BIBentryALTinterwordspacing
M.~A. Riedmiller, R.~Hafner, T.~Lampe, M.~Neunert, J.~Degrave, T.~V. de~Wiele,
  V.~Mnih, N.~Heess, and J.~T. Springenberg, ``Learning by playing - solving
  sparse reward tasks from scratch,'' \emph{CoRR}, vol. abs/1802.10567, 2018.
  [Online]. Available: \url{http://arxiv.org/abs/1802.10567}
\BIBentrySTDinterwordspacing

\bibitem{wulfmeiercompositional}
\BIBentryALTinterwordspacing
M.~Wulfmeier, A.~Abdolmaleki, R.~Hafner, J.~T. Springenberg, M.~Neunert,
  T.~Hertweck, T.~Lampe, N.~Y. Siegel, N.~Heess, and M.~A. Riedmiller,
  ``Regularized hierarchical policies for compositional transfer in robotics,''
  \emph{CoRR}, vol. abs/1906.11228, 2019. [Online]. Available:
  \url{http://arxiv.org/abs/1906.11228}
\BIBentrySTDinterwordspacing

\bibitem{deisenroth2014mtrl}
\BIBentryALTinterwordspacing
M.~P. Deisenroth, P.~Englert, J.~Peters, and D.~Fox, ``Multi-task policy search
  for robotics,'' in \emph{2014 {IEEE} International Conference on Robotics and
  Automation, {ICRA} 2014, Hong Kong, China, May 31 - June 7, 2014}.\hskip 1em
  plus 0.5em minus 0.4em\relax {IEEE}, 2014, pp. 3876--3881. [Online].
  Available: \url{https://doi.org/10.1109/ICRA.2014.6907421}
\BIBentrySTDinterwordspacing

\bibitem{suttonandbarto}
R.~S. Sutton and A.~G. Barto, \emph{Reinforcement learning: An introduction},
  2018.

\bibitem{konda}
\BIBentryALTinterwordspacing
V.~R. Konda and J.~N. Tsitsiklis, ``Actor-critic algorithms,'' in
  \emph{Advances in Neural Information Processing Systems 12, {[NIPS}
  Conference, Denver, Colorado, USA, November 29 - December 4, 1999]}, S.~A.
  Solla, T.~K. Leen, and K.~M{\"{u}}ller, Eds.\hskip 1em plus 0.5em minus
  0.4em\relax The {MIT} Press, 1999, pp. 1008--1014. [Online]. Available:
  \url{http://papers.nips.cc/paper/1786-actor-critic-algorithms}
\BIBentrySTDinterwordspacing

\bibitem{eysenbachsmm}
\BIBentryALTinterwordspacing
L.~Lee, B.~Eysenbach, E.~Parisotto, E.~P. Xing, S.~Levine, and
  R.~Salakhutdinov, ``Efficient exploration via state marginal matching,''
  \emph{CoRR}, vol. abs/1906.05274, 2019. [Online]. Available:
  \url{http://arxiv.org/abs/1906.05274}
\BIBentrySTDinterwordspacing

\bibitem{haarnoja2018soft}
T.~Haarnoja, A.~Zhou, K.~Hartikainen, G.~Tucker, S.~Ha, J.~Tan, V.~Kumar,
  H.~Zhu, A.~Gupta, P.~Abbeel \emph{et~al.}, ``Soft actor-critic algorithms and
  applications,'' \emph{arXiv preprint arXiv:1812.05905}, 2018.

\bibitem{RND}
\BIBentryALTinterwordspacing
Y.~Burda, H.~Edwards, A.~J. Storkey, and O.~Klimov, ``Exploration by random
  network distillation,'' in \emph{7th International Conference on Learning
  Representations, {ICLR} 2019, New Orleans, LA, USA, May 6-9, 2019}.\hskip 1em
  plus 0.5em minus 0.4em\relax OpenReview.net, 2019. [Online]. Available:
  \url{https://openreview.net/forum?id=H1lJJnR5Ym}
\BIBentrySTDinterwordspacing

\end{thebibliography}

\newpage
\section*{Appendix A. Reward Functions and Additional Task Details}
\label{appendix:a}

\subsection{In-Hand Manipulation Tasks on Hardware}
The rewards for this family of tasks are defined as follows. $\theta^{x}$ represent the Sawyer end-effector's wrist euler angle, $x, y, z$ represent the object's 3D position, and $\hat\theta^{z}$ represents the object's z euler angle. $x_{goal}$ and others represent the task's goal position or angle. The threshold for determining whether the object has been lifted is subjected to the real world arena size. We set it to 0.1m in our experiment.

\begin{align*}
R_{recenter} = - 3||\begin{bmatrix}x \\ y\end{bmatrix} - \begin{bmatrix}x_{goal} \\ y_{goal}\end{bmatrix} ||
    - ||\begin{bmatrix}x \\ y \\ z\end{bmatrix} - \begin{bmatrix}x_{hand} \\ y_{hand} \\ z_{hand}\end{bmatrix} ||
\end{align*}

\begin{align*}R_{lift} = - |z-z_{goal}|
\end{align*}

\begin{align*}
    R_{flipup} = &-5 | \theta^{x} -  \theta^{x}_{goal}| -50 (\mathbb{1} \Bigl\{ z < \text{threshold}\Bigr\} ) + \\ 
    &10 (\mathbb{1} \Bigl\{ | \theta^{x} -  \theta^{x}_{goal}| < 0.15 \text{ AND } z > \text{threshold} \Bigr\} )
\end{align*}

\begin{align*}R_{reorient} = -|\hat\theta^{z}-\hat\theta^{z}_{goal}|
\end{align*}

A rollout is considered a success (as reported in the figures) if it reaches a state where the valve is in-hand and flipped facing up:
\begin{align*}
    z > \text{threshold} \text{ AND } |\theta^{x} - \theta^{x}_{goal}| < 0.15
\end{align*}

\subsection{Pipe Insertion Tasks on Hardware}
The rewards for this family of tasks are defined as follows. 
$x, y, z$ represent the object's 3D position, and $q$ represent the joint positions of the D'Hand. $x_{goal}$ and others represent the task's goal position or angle. The threshold for determining whether the object has been lifted is subjected to the real world arena size. We set it to 0.1m in our experiment. To reduce collision in real world experiments, we have two tasks for insertion. One approaches the peg and the other attempts insertion.

\begin{align*}
R_{recenter} = - 3||\begin{bmatrix}x \\ y\end{bmatrix} - \begin{bmatrix}x_{goal} \\ y_{goal}\end{bmatrix} ||
    - ||\begin{bmatrix}x \\ y \\ z\end{bmatrix} - \begin{bmatrix}x_{hand} \\ y_{hand} \\ z_{hand}\end{bmatrix} ||
\end{align*}

\begin{align*}R_{lift} = - &2 |z-z_{goal}| - 2|q-q_{goal}|
\end{align*}

\begin{align*}R_{Insert1} = -d_1 + 10(\mathbb{1} \Bigl\{d_1<0.1\Bigr\})
\end{align*}

\begin{align*}R_{Insert2} = -d_2 + 10(\mathbb{1} \Bigl\{d_2<0.1\Bigr\})
\end{align*}

$where$
\begin{align*}d_1 = ||\begin{bmatrix}x \\ y \\ z\end{bmatrix} - \begin{bmatrix}x_{goal1} \\ y_{goal1} \\ z_{goal1}\end{bmatrix} ||
\end{align*}
\begin{align*}d_2 = ||\begin{bmatrix}x \\ y \\ z\end{bmatrix} - \begin{bmatrix}x_{goal2} \\ y_{goal2} \\ z_{goal2}\end{bmatrix} ||
\end{align*}

\begin{align*}R_{Remove} = -||\begin{bmatrix}x \\ y \\ z\end{bmatrix} - \begin{bmatrix}x_{arena\_center} \\ y_{arena\_center} \\ z_{arena\_center}\end{bmatrix} ||
\end{align*}

A rollout is considered a success (as reported in the figures) if it reaches a state where the valve is in-hand and pipe is inserted:
\begin{align*}
    z > \text{threshold} \text{ AND } d_2 < 0.05
\end{align*}

\subsection{Lightbulb Insertion Tasks in Simulation}

The rewards for this family of tasks include $R_{recenter}, R_{pickup}, R_{flipup}$ defined in the previous section, as well as the following bulb insertion reward:

\begin{align*}
    R_{bulb} = &-||\begin{bmatrix}x \\ y\end{bmatrix} - \begin{bmatrix}x_{goal} \\ y_{goal}\end{bmatrix} || - 2(|z - z_{goal}|)
    + \\ &\mathbb{1} \Bigl\{ ||\begin{bmatrix}x \\ y\end{bmatrix} - \begin{bmatrix}x_{goal} \\ y_{goal}\end{bmatrix} || < 0.1 \Bigr\} \\
    + \\ &10 ( \mathbb{1} \Bigl\{ ||\begin{bmatrix}x \\ y\end{bmatrix} - \begin{bmatrix}x_{goal} \\ y_{goal}\end{bmatrix} || < 0.1 \text{ AND } |z - z_{goal}| < 0.1 \Bigr\} )
    -\\ &\mathbb{1} \Bigl\{ z < \text{threshold} \Bigr\}
\end{align*}

A rollout is considered a success (as reported in the figures) if it reaches a state where the bulb is positioned very close to the goal position in the lamp:
\begin{align*}
    || \begin{bmatrix}x \\ y\end{bmatrix} - \begin{bmatrix}x_{goal} \\ y_{goal}\end{bmatrix} || < 0.1 \text{ AND } |z - z_{goal}| < 0.1
\end{align*}

\subsection{Basketball Tasks in Simulation}

The rewards for this family of tasks include $R_{recenter}, R_{pickup}$ defined in the previous section, as well as the following basket dunking reward:

\begin{align*}
    R_{basket} = &-||\begin{bmatrix}x \\ y \\ z\end{bmatrix} - \begin{bmatrix}x_{goal} \\ y_{goal} \\ z_{goal}\end{bmatrix} ||
    + \\ & 20 (\mathbb{1} \Bigl\{ ||\begin{bmatrix}x \\ y \\ z\end{bmatrix} - \begin{bmatrix}x_{goal} \\ y_{goal} \\ z_{goal}\end{bmatrix} || < 0.2 \Bigr\}) \\
    &+ 50 (\mathbb{1} \Bigl\{ ||\begin{bmatrix}x \\ y \\ z\end{bmatrix} - \begin{bmatrix}x_{goal} \\ y_{goal} \\ z_{goal}\end{bmatrix} || < 0.1 \Bigr\} )
    - \\ &\mathbb{1} \Bigl\{ z < \text{threshold} \Bigr\}
\end{align*}

A rollout is considered a success (as reported in the figures) if it reaches a state where the ball is positioned very close to the goal position above the basket:
\begin{align*}
    || \begin{bmatrix}x \\ y\end{bmatrix} - \begin{bmatrix}x_{goal} \\ y_{goal}\end{bmatrix} || < 0.1 \text{ AND } |z - z_{goal}| < 0.15
\end{align*}

\section*{Appendix B. Additional Domains}
\label{appendix:b}

\begin{figure}[!h]
  \centering
  \hspace{0.2cm}
  \includegraphics[width= .3\textwidth]{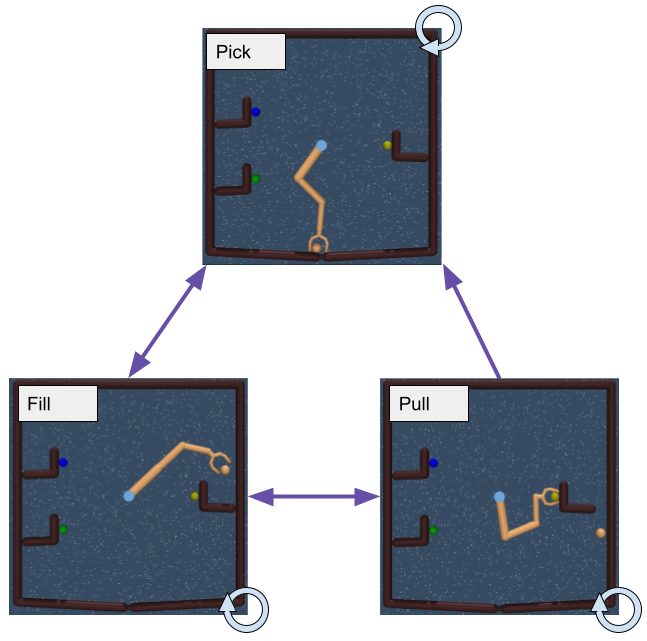}
  \\ \vspace{0.5cm}
  \includegraphics[width= 0.45\columnwidth]{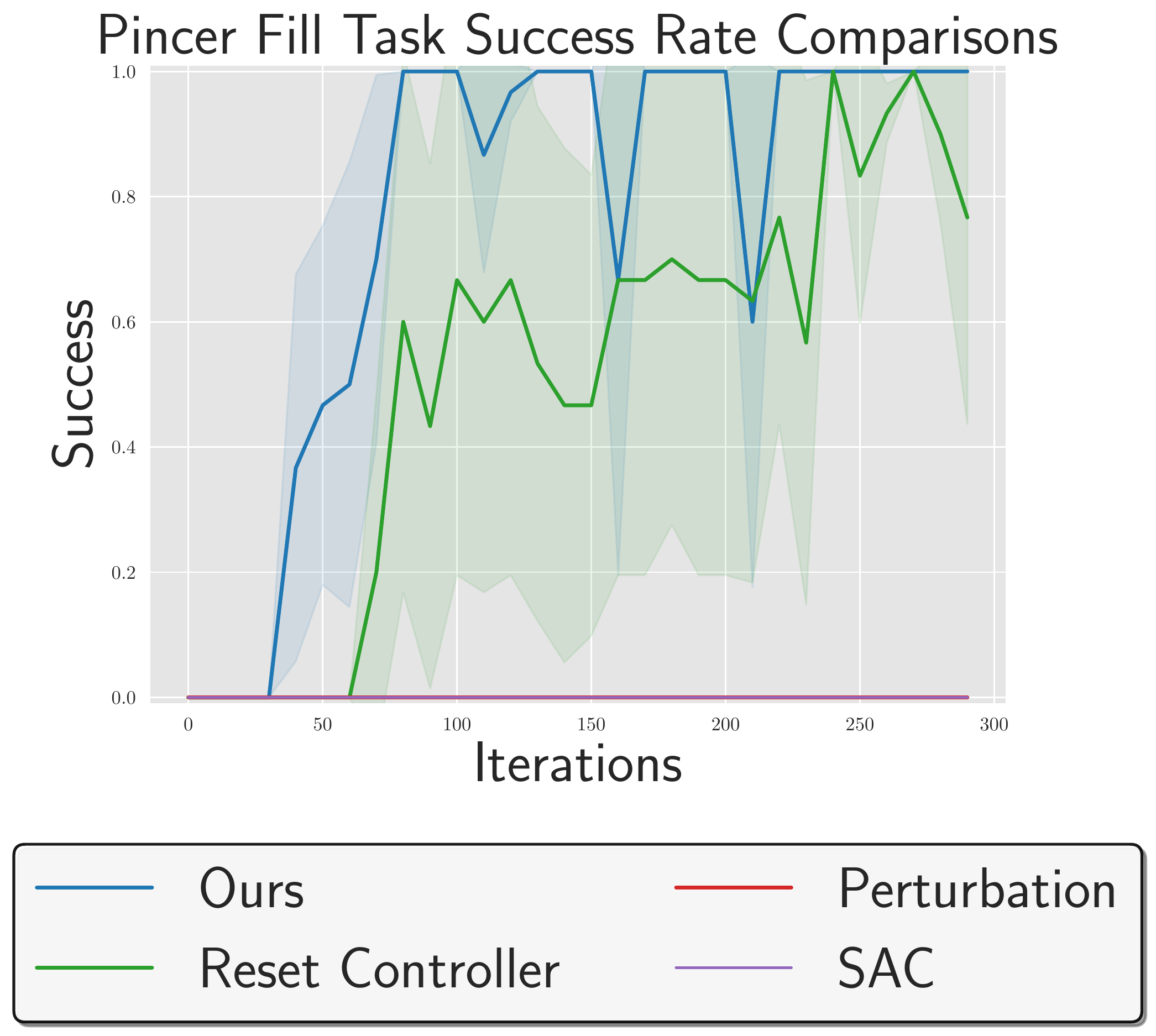}
  \includegraphics[width= 0.41\columnwidth]{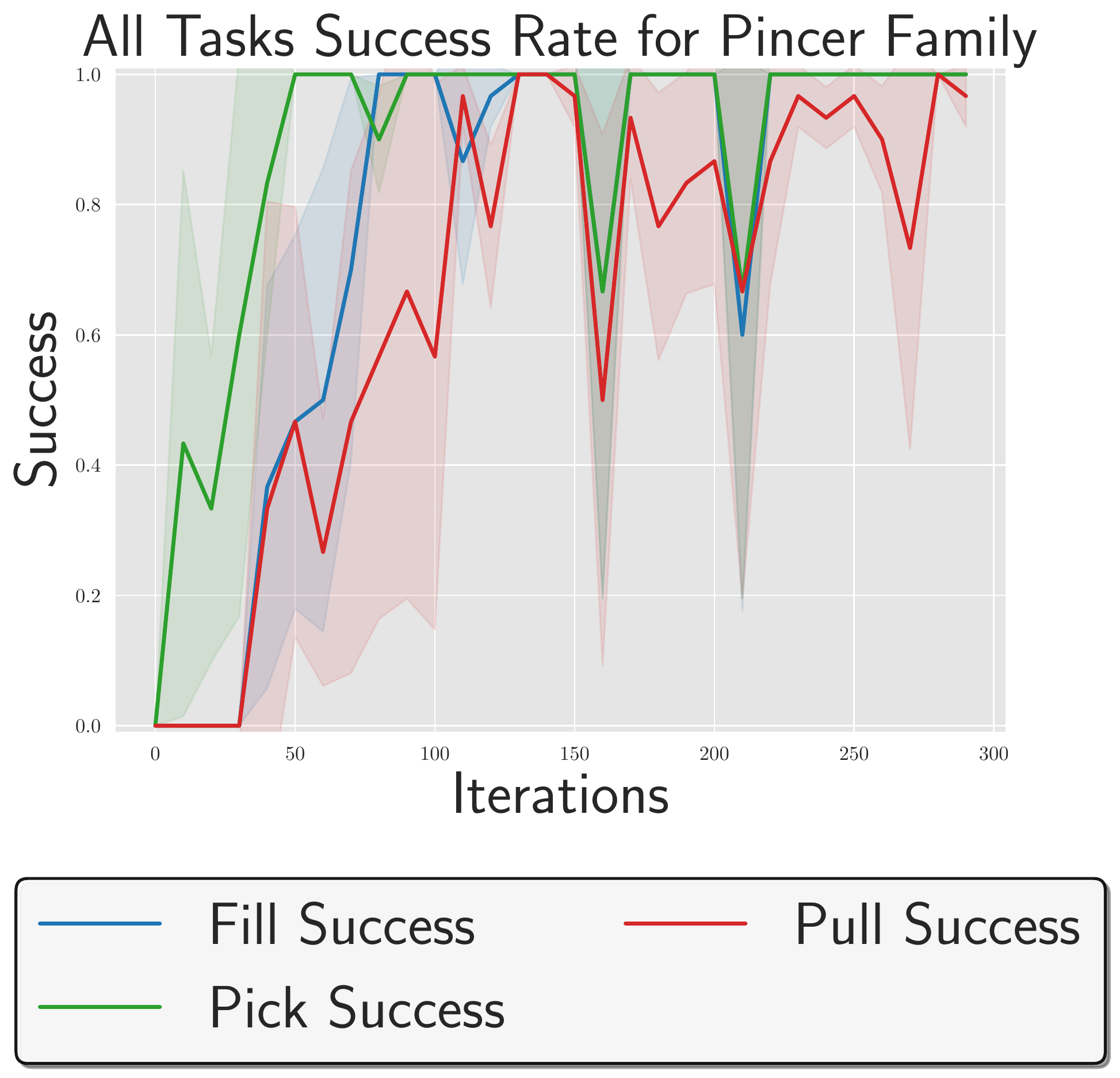}
  \caption{Pincer domain - object grasping, filling the drawer with the object, pulling open the drawer. These tasks naturally form a cycle - once an object is picked up, it can be filled in the drawer, following which the drawer can be pulled open and grasping and filling can be practiced again.}
  \label{fig:Pincer-Drawer-Domain}
\end{figure}

In addition to the test domains described in Section \ref{sec:sim-setup}, we also tested our method in simulation on simpler tasks such as a 2D ``pincer'' and a simplified lifting task on the Sawyer and ``D'Hand'' setup. The pincer task is described in Figure \ref{fig:Pincer-Drawer-Domain}. Figure \ref{fig:Pincer-Drawer-Domain} shows the performance of our method as well as baseline comparisons.

\section*{Appendix C. Hyperparameter Details}
\label{appendix:c}

\begin{table}[!h]
    \centering
    \begin{tabular}{| p{3cm}||p{4cm} |}
     \hline
     \textbf{SAC} & \\
     \hline
     Learning Rate & $3 \times 10^{-4}$\\
     Discount Factor $\gamma$ & $0.99$\\
     Policy Type & Gaussian\\
     Policy Hidden Sizes & $(512, 512)$\\
     RL Batch Size & $1024$ \\
     Reward Scaling & $1$\\
     Replay Buffer Size & $500,000$\\
     Q Hidden Sizes & $(512, 512)$\\
     Q Hidden Activation & ReLU\\
     Q Weight Decay & $0$ \\
     Q Learning Rate & $3 \times 10^{-4}$\\
     Target Network $\tau$ & $5\times10^{-3}$ \\
     \hline
    \end{tabular}
\caption{Hyperparameters used across all domains.}
\end{table}

\section*{Appendix D. Task Graph Details}
\label{appendix:d}
We provide some details of the task graphs for every domain below.

\begin{algorithm}[!h]
\begin{algorithmic}[1]
\REQUIRE Euclidean coordinates of object $q$, Sawyer wrist angle $\theta$, previous task $\phi$
\STATE $is\_lifted = q_z > 0.15$
\STATE $is\_upright = |\theta - \theta_{upright}| < 0.1$
\STATE $not\_centered = |q - q_{center}| > 0.1$

\IF{$is\_upright$ and $is\_lifted$}
\RETURN $Inhand$
\ELSIF{$is\_lifted$}
\RETURN $Flipup$
\ELSIF{$not\_centered$ and $\phi=Recenter$}
\RETURN $Perturb$
\ELSIF{$not\_centered$}
\RETURN $Recenter$
\ELSE
\RETURN $Lift$
\ENDIF
\end{algorithmic}
\caption{{\bf In-Hand Manipulation Task Graph (Hardware)}}
\label{alg:task_graph_1}
\end{algorithm}

\begin{algorithm}[!h]
\begin{algorithmic}[1]
\REQUIRE Euclidean coordinates of object $q$, a waypoint close to peg $q_{waypoint}$, previous task $\phi$
\STATE $is\_lifted = q_z > 0.15$
\STATE $is\_inserted = |q - q_{inserted}| < 0.05$
\STATE $close\_to\_waypoint = |q - q_{waypoint}| < 0.05$
\STATE $not\_centered = |q - q_{center}| > 0.1$

\IF{$is\_inserted$}
\RETURN $Remove$
\ELSIF{$close\_to\_waypoint$}
\RETURN $Insert2$
\ELSIF{$is\_lifted$}
\RETURN $Insert1$
\ELSIF{$not\_centered$ and $\phi=Recenter$}
\RETURN $Perturb$
\ELSIF{$not\_centered$}
\RETURN $Recenter$
\ELSE
\RETURN $Lift$
\ENDIF
\end{algorithmic}
\caption{{\bf Pipe Insertion Task Graph (Hardware)}}
\label{alg:task_graph_2}
\end{algorithm}

\begin{algorithm}[!h]
\begin{algorithmic}[1]
\REQUIRE Object position $\begin{bmatrix}x \\ y \\ z\end{bmatrix}$, Sawyer wrist angle (its $x$ Euler angle) $\theta^x$, previous task $\phi$
\STATE Let $\begin{bmatrix}x_{center} \\ y_{center}\end{bmatrix}$ be the center coordinates of the arena (relative to the Sawyer base).
\STATE Let $z_{threshold}$ be the height (in meters) above the arena that we consider the object to be ``picked up.''
\STATE $is\_centered = ||\begin{bmatrix}x \\ y\end{bmatrix} - \begin{bmatrix}x_{center} \\ y_{center}\end{bmatrix}|| < 0.1$
\STATE $is\_lifted = z > z_{threshold}$
\STATE $is\_facing\_up = |\theta^x - \theta^x_{upright}| < 0.1$

\IF{NOT $is\_centered$ and NOT $is\_lifted$}
\IF{$\phi = \text{Recenter}$}
\RETURN Perturb
\ELSE
\RETURN Recenter
\ENDIF
\ELSIF{$is\_centered$ and NOT $is\_lifted$}
\RETURN Lift
\ELSIF{$is\_lifted$ and NOT $is\_facing\_up$}
\RETURN Flip Up
\ELSIF{$is\_lifted$ and $is\_facing\_up$}
\RETURN Lightbulb Insertion
\ENDIF
\end{algorithmic}

\caption{{\bf Lightbulb Insertion Task Graph (Simulation)}}
\label{alg:task_graph_3}
\end{algorithm}

\begin{algorithm}[!h]
\begin{algorithmic}[1]
\REQUIRE Object position $\begin{bmatrix}x \\ y \\ z\end{bmatrix}$, previous task $\phi$
\STATE Let $\begin{bmatrix}x_{center} \\ y_{center}\end{bmatrix}$ be the coordinates of the arena where we want to pick up the ball, such that it is out of the way of the hoop (relative to the Sawyer base).
\STATE Let $\theta^x_{upright}$ be the wrist angle (in radians) that we want the hand to be facing. ($\theta_{upright}^x = \pi$ in our instantiation).
\STATE Let $z_{threshold}$ be the height (in meters) above the arena that we consider the object to be ``picked up.''
\STATE $is\_centered = ||\begin{bmatrix}x \\ y\end{bmatrix} - \begin{bmatrix}x_{center} \\ y_{center}\end{bmatrix}|| < 0.1$
\STATE $is\_lifted = z > z_{threshold}$

\IF{NOT $is\_centered$ and NOT $is\_lifted$}
\IF{$\phi = \text{Recenter}$}
\RETURN Perturb
\ELSE
\RETURN Recenter
\ENDIF
\ELSIF{$is\_centered$ and NOT $is\_lifted$}
\RETURN Lift
\ELSIF{$is\_lifted$}
\RETURN Basketball Dunking
\ENDIF
\end{algorithmic}
\caption{{\bf Basketball Task Graph (Simulation)}}
\label{alg:task_graph_4}
\end{algorithm}

\end{document}